\documentclass[conference]{IEEEtran}
\IEEEoverridecommandlockouts
\usepackage{cite}
\usepackage{amsmath,amssymb,amsfonts}
\usepackage{algorithmic}
\usepackage{graphicx}
\usepackage{textcomp}
\usepackage{xcolor}
\usepackage{graphicx}
\usepackage{subfig}
\usepackage{multirow}
\usepackage{url}

\def\BibTeX{{\rm B\kern-.05em{\sc i\kern-.025em b}\kern-.08em
    T\kern-.1667em\lower.7ex\hbox{E}\kern-.125emX}}
\begin{document}

\title{NutritionVerse: Empirical Study of Various Dietary Intake Estimation Approaches}

\author{\IEEEauthorblockN{Chi-en Amy Tai}
\IEEEauthorblockA{\textit{Systems Design Engineering} \\
\textit{University of Waterloo}\\
Waterloo, ON \\
0000-0001-7023-8784}
\and
\IEEEauthorblockN{Matthew Keller}
\IEEEauthorblockA{\textit{Systems Design Engineering} \\
\textit{University of Waterloo}\\
Waterloo, ON \\
0009-0008-0510-6628}
\and
\IEEEauthorblockN{Saeejith Nair}
\IEEEauthorblockA{\textit{Systems Design Engineering} \\
\textit{University of Waterloo}\\
Waterloo, ON \\
0000-0003-2488-5141}
\and
\IEEEauthorblockN{Yuhao Chen}
\IEEEauthorblockA{\textit{Systems Design Engineering} \\
\textit{University of Waterloo}\\
Waterloo, ON \\
0000-0001-6094-0545}
\and
\IEEEauthorblockN{Yifan Wu}
\IEEEauthorblockA{\textit{Systems Design Engineering} \\
\textit{University of Waterloo}\\
Waterloo, ON \\
0000-0002-2856-9835}
\and
\IEEEauthorblockN{Olivia Markham}
\IEEEauthorblockA{\textit{Systems Design Engineering} \\
\textit{University of Waterloo}\\
Waterloo, ON \\
0009-0005-7915-1024}
\and
\IEEEauthorblockN{Krish Parmar}
\IEEEauthorblockA{\textit{Systems Design Engineering} \\
\textit{University of Waterloo}\\
Waterloo, ON \\
0009-0004-4191-8835}
\and
\IEEEauthorblockN{Pengcheng Xi}
\IEEEauthorblockA{\textit{National Research Council Canada}\\
Ottawa, ON \\
0000-0003-3236-5234}
\and
\IEEEauthorblockN{Heather Keller}
\IEEEauthorblockA{\textit{Kinesiology and Health Sciences} \\
\textit{University of Waterloo}\\
Waterloo, ON \\
0000-0001-7782-8103}
\and
\IEEEauthorblockN{Sharon Kirkpatrick}
\IEEEauthorblockA{\textit{School of Public Health Sciences} \\
\textit{University of Waterloo}\\
Waterloo, ON \\
0000-0001-9896-5975}
\and
\IEEEauthorblockN{Alexander Wong}
\IEEEauthorblockA{\textit{Systems Design Engineering} \\
\textit{University of Waterloo}\\
Waterloo, ON \\
0000-0002-5295-2797}
}

\maketitle

\begin{abstract}
Accurate dietary intake estimation is critical for informing policies and programs to support healthy eating, as malnutrition has been directly linked to decreased quality of life. However self-reporting methods such as food diaries suffer from substantial bias. Other conventional dietary assessment techniques and emerging alternative approaches such as mobile applications incur high time costs and may necessitate trained personnel. Recent work has focused on using computer vision and machine learning to automatically estimate dietary intake from food images, but the lack of comprehensive datasets with diverse viewpoints, modalities and food annotations hinders the accuracy and realism of such methods. To address this limitation, we introduce NutritionVerse-Synth, the first large-scale dataset of 84,984 photorealistic synthetic 2D food images with associated dietary information and multimodal annotations (including depth images, instance masks, and semantic masks). Additionally, we collect a real image dataset, NutritionVerse-Real, containing 889 images of 251 dishes to evaluate realism. Leveraging these novel datasets, we develop and benchmark NutritionVerse, an empirical study of various dietary intake estimation approaches, including indirect segmentation-based and direct prediction networks. We further fine-tune models pretrained on synthetic data with real images to provide insights into the fusion of synthetic and real data. Finally, we release both datasets (NutritionVerse-Synth, NutritionVerse-Real) on \url{https://www.kaggle.com/nutritionverse/datasets} as part of an open initiative to accelerate machine learning for dietary sensing.
\end{abstract}


\begin{figure*}
    \captionsetup[subfigure]{labelformat=empty, position=top, labelfont=bf}
    \centering
    \subfloat[RGB Image]{\frame{\includegraphics[width=.245\linewidth]{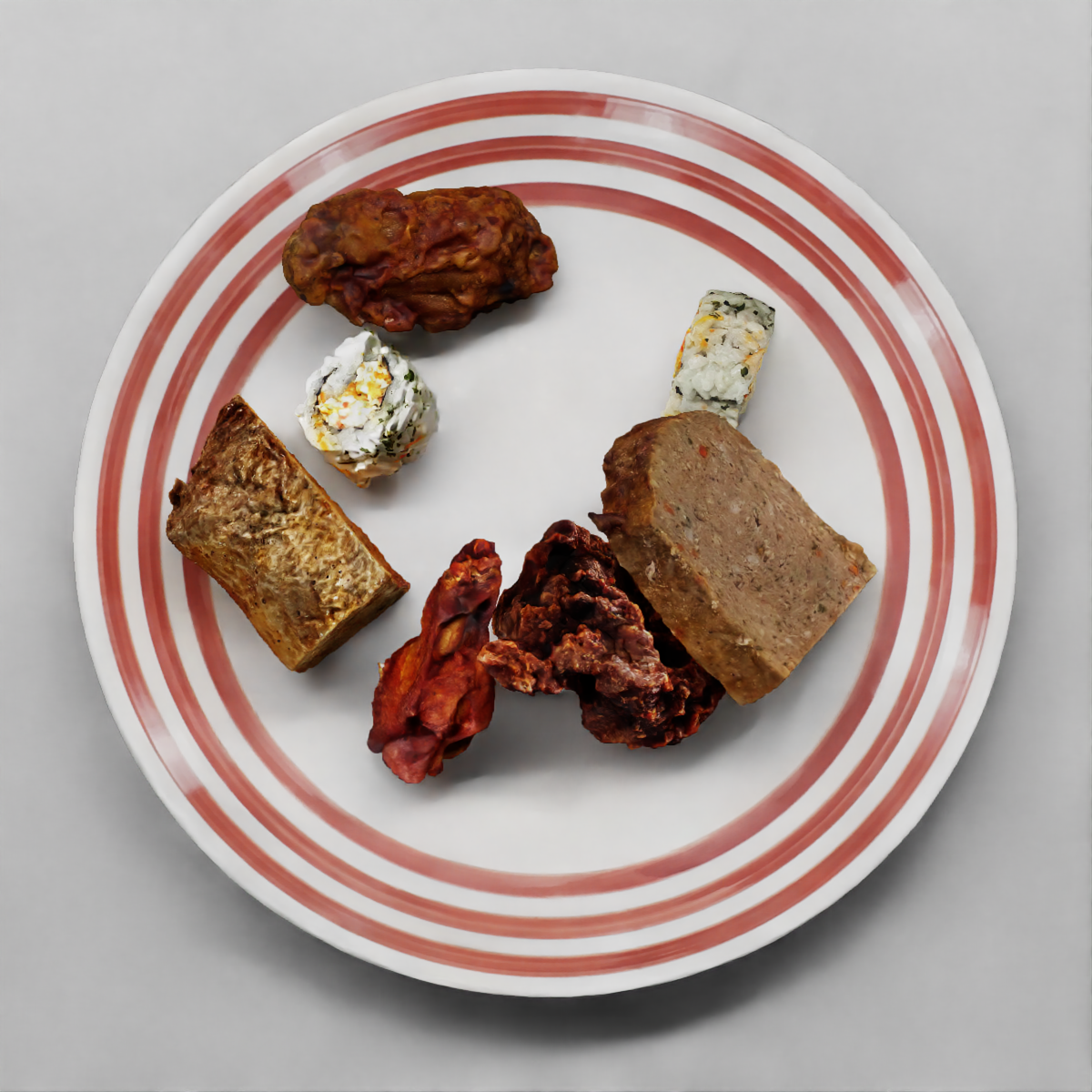}}}
    \subfloat[Depth Image]{\frame{\includegraphics[width=.245\linewidth]{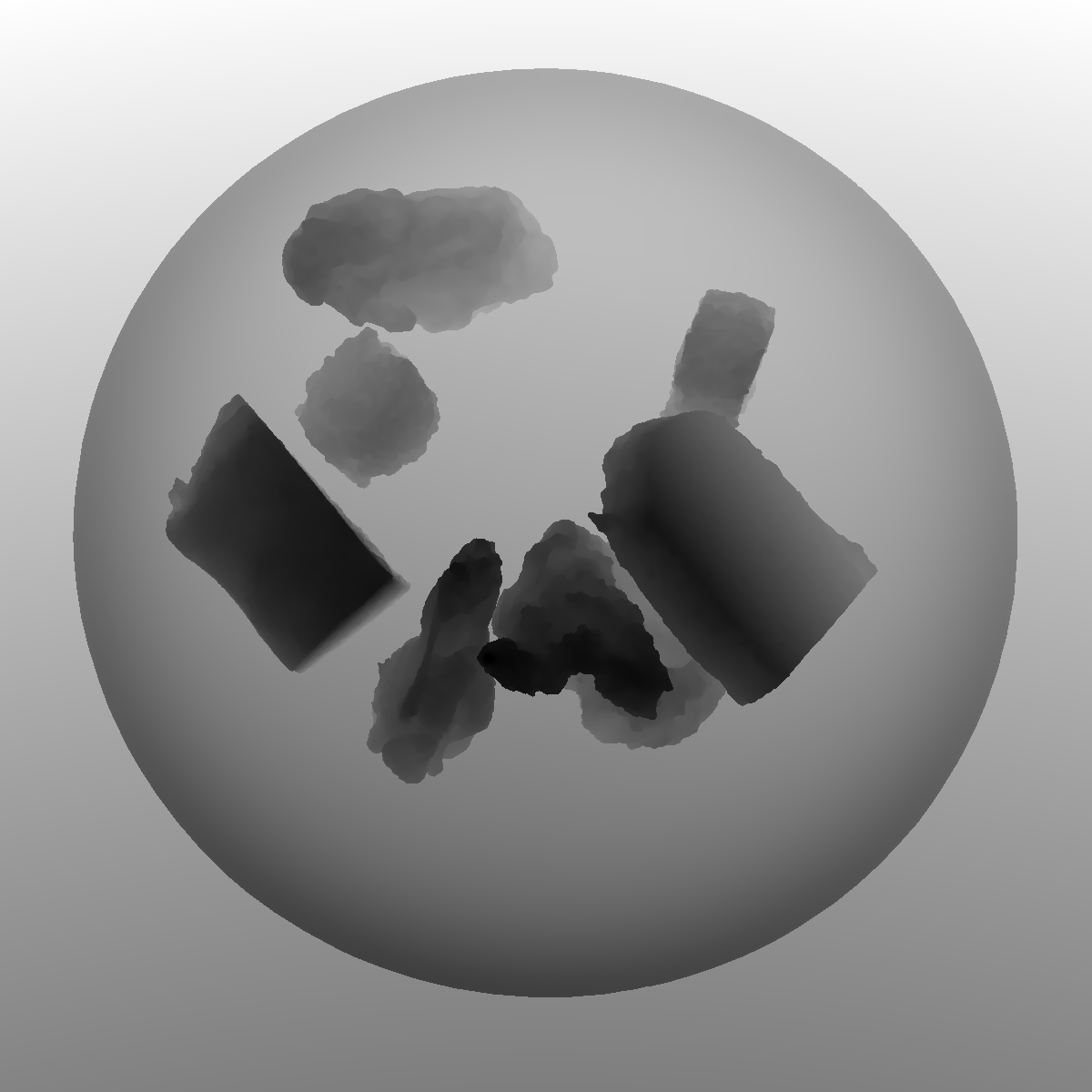}}}
    \subfloat[Instance Segmentation]{\frame{\includegraphics[width=.245\linewidth]{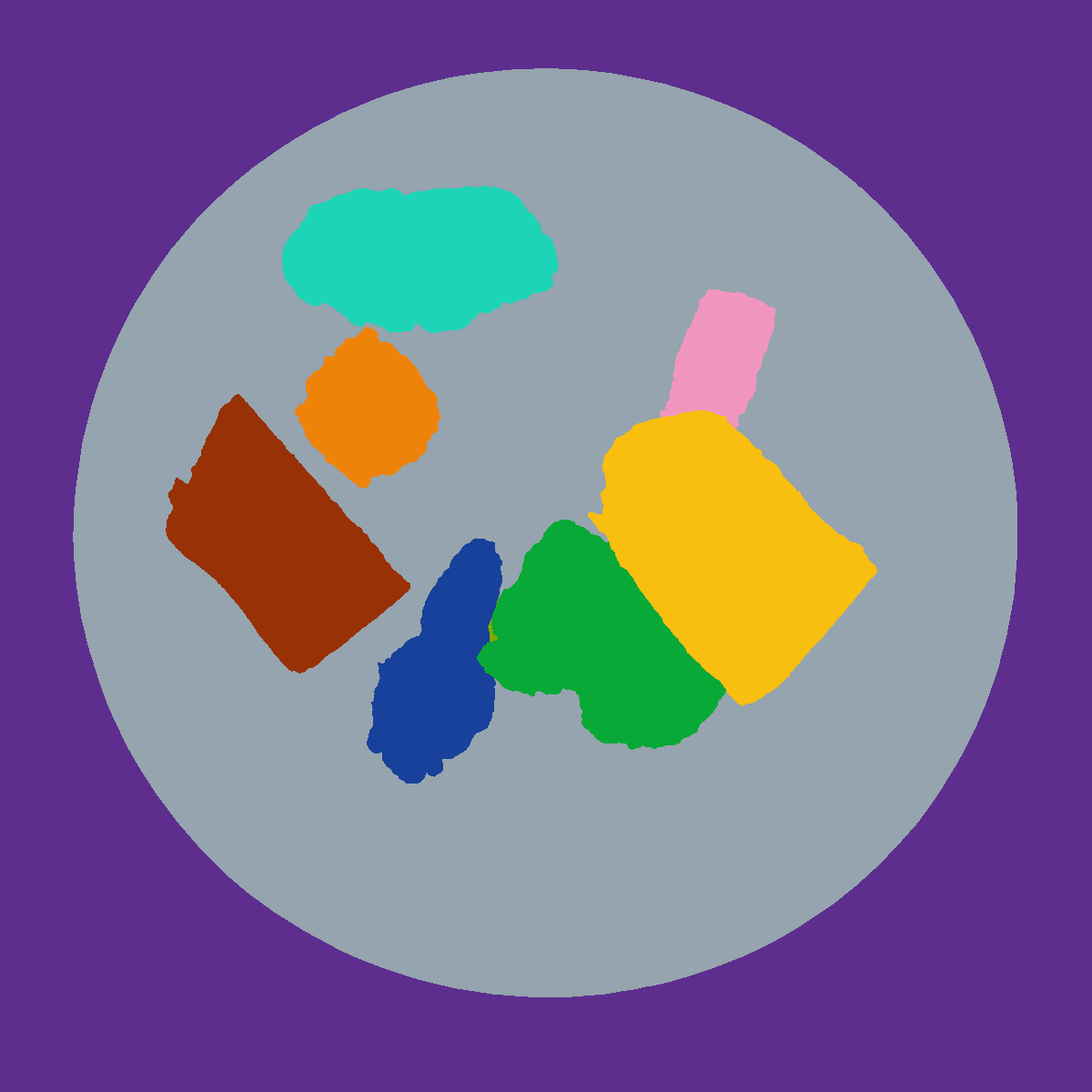}}}
    \subfloat[Semantic Segmentation]{\frame{\includegraphics[width=.245\linewidth]{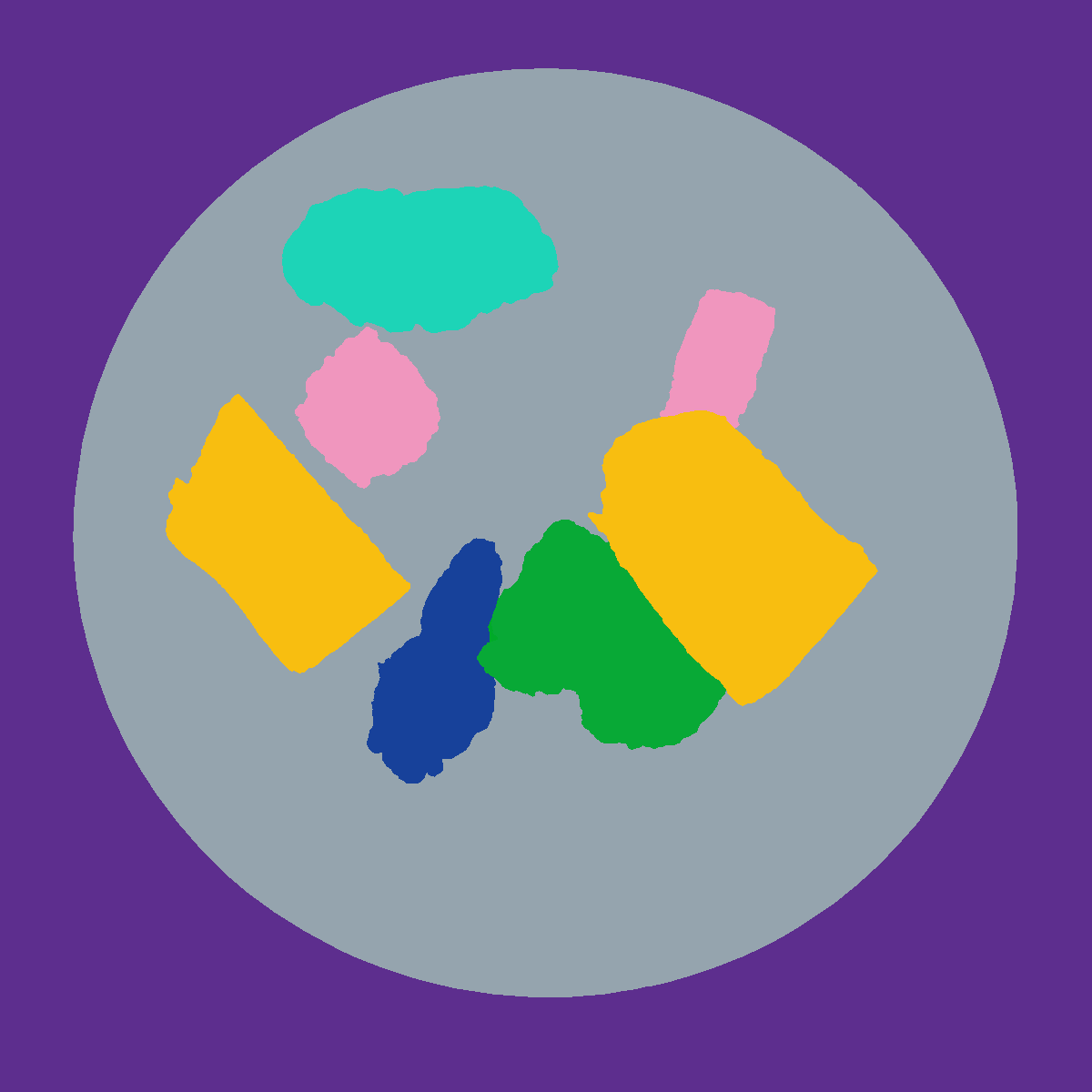}}}
  \caption{Sample scene from NV-Synth dataset with the associated multi-modal image data (e.g., RGB and depth data) and annotation metadata (e.g., instance and semantic segmentation masks) derived using objects from the NutritionVerse-3D dataset~\cite{nutritionverse-3d}. There are 2 meatloaves, 1 chicken leg, 1 chicken wing, 1 pork rib, and 2 sushi rolls in this scene.}
    \label{fig:process}
\end{figure*}

\maketitle

\section{Introduction}
\begin{table*}[!ht]
 \centering
\begin{tabular}{llrrlcclccllll}
\hline
\multirow{2}{*}{\textbf{Work}} & \multirow{2}{*}{\textbf{Public}} & \multicolumn{7}{c}{\textbf{Data}} & \multicolumn{5}{c}{\textbf{Dietary Info}} \\ \cline{3-14} 
 & & \multicolumn{1}{l}{\textbf{\# Img}} & \multicolumn{1}{l}{\textbf{\# Items}} & \textbf{Real} & \textbf{Mixed} & \textbf{\# Angles} & \textbf{Depth} & \multicolumn{1}{c}{\textbf{Annotation Masks}} & \multicolumn{1}{l}{\textbf{CL}} & \textbf{M} & \textbf{P} & \textbf{F} & \textbf{CB} \\ \hline
\cite{10.1145/3347448.3357172} & \checkmark & 18 & 3 & Y & N & 1 &  &  & \checkmark &  &  &  &  \\
\cite{7045971} & \checkmark & 646 & 41 & Y & Y & 1 &  &  &  \checkmark &  &  &  &  \\
\cite{7410503} & \checkmark & 50,374 & 201 & Y & Y & 1 &  & &  \checkmark &  &  &  &  \\
\cite{2017arXiv170507632L} & \checkmark & 2,978 & 160 & Y & N & 2 &  &  & \checkmark & \multicolumn{1}{c}{\checkmark} &  &  &  \\
\cite{thames2021nutrition5k} & \checkmark & 5,006 & 555 & Y & Y & 4 & \multicolumn{1}{c}{\checkmark} &  & \checkmark & \multicolumn{1}{c}{\checkmark} & \multicolumn{1}{c}{\checkmark} & \multicolumn{1}{c}{\checkmark} & \multicolumn{1}{c}{\checkmark} \\
\cite{6748066} &  & 3000 & 8 & Y & Y & 2 & & \checkmark & \checkmark & \checkmark & & & \\
NV-Real & \checkmark & 889 & 45 & Y & Y & 4 &  & \checkmark & \checkmark & \multicolumn{1}{c}{\checkmark} & \multicolumn{1}{c}{\checkmark} & \multicolumn{1}{c}{\checkmark} & \multicolumn{1}{c}{\checkmark} \\
NV-Synth & \checkmark & 84,984 & 45 & N & Y & 12 & \multicolumn{1}{c}{\checkmark} & \checkmark & \checkmark & \multicolumn{1}{c}{\checkmark} & \multicolumn{1}{c}{\checkmark} & \multicolumn{1}{c}{\checkmark} & \multicolumn{1}{c}{\checkmark}
\end{tabular}
\caption{Overview of existing dietary intake estimation datasets compared to ours where Mixed refers to whether multiple food item types are present in an image, and CL refers to calories, M to mass, P to protein, F to fat, and CB to carbohydrate.}
\label{tab:research-overview}
\end{table*}

Accurate dietary intake estimation is critical for informing policies and programs to support healthy eating, as malnutrition has been directly linked to decreased quality of life ~\cite{malnutrition-qol}. However, conventional diet assessment techniques such as food frequency questionnaires, food diaries, and 24-hour recall ~\cite{automated-dietary-recall} are subject to substantial bias ~\cite{kipnis2003structure,freedman2014pooled,freedman2015pooled}. Emerging alternative approaches for diet assessment, including mobile applications~\cite{mobile-phone-applications, snap-and-eat}, digital photography~\cite{digital-photography}, and personal assistants~\cite{personal-assistant} incur high time costs and may necessitate trained personnel. Fortunately, recent promising methods combine these alternative methods with computer vision and machine learning algorithms to automatically estimate nutritional information from food images~\cite{food-recog-promise, 10.1145/3347448.3357172}. 

Existing literature~\cite{10.1145/3347448.3357172,7045971,7410503,2017arXiv170507632L,thames2021nutrition5k} collects images of real scenes to train models that achieve high accuracy. However, these techniques operate on fixed modalities and viewpoints, hindering systematic comparison due to data limitations. For example,~\cite{7045971} is only trained and evaluated on the RGB image of the top view of a food scene. Furthermore, current food recognition and intake estimation methods face several key limitations: restricted output variables (e.g. only calories or mass), lack of diverse viewpoints or incomplete food annotations in datasets, and biases from predefined camera angles during data capture. 

Subsequently, the lack of a comprehensive high-quality image dataset hinders the accuracy and realism of systems based on machine learning and computer vision. For such dietary intake estimation systems to be effective, diverse high-quality training data capturing multiple angles and modalities are required. However, manual creation of large-scale datasets with such diversity is time-consuming and hard to scale. On the other hand, synthesized 3D food models enable view augmentation to generate countless photorealistic 2D renderings from any viewpoint, reducing imbalance across camera angles. As shown in Figure~\ref{fig:process}, leveraging 3D assets facilitates creation of rich multi-modal datasets (e.g., RGB, depth) with photorealistic images, perfect annotations, and dietary metadata through algorithmic scene composition. Compared to existing datasets that are focused solely on quantity, our contributions also address the gap in the quality of the data by procedurally generating scenes that span a huge diversity of food items, placements, and camera angles.

In this paper, we present a process to collect a large image dataset of food scenes that span diverse viewpoints. We first leverage high-quality photorealistic 3D food models and introduce NutritionVerse-Synth (NV-Synth), a dataset of 84,984 high-resolution 2D food images algorithmically rendered from 7,081 unique scenes, along with associated diet information derived from the 3D models. To evaluate realism, we also collect the NutritionVerse-Real (NV-Real) dataset of 889 manually captured images across 251 distinct dishes. We benchmark various intake estimation approaches on these datasets and present NutritionVerse, a collection of models that estimate intake from 2D food images. We release both the synthetic and real-world datasets at \url{https://www.kaggle.com/nutritionverse/datasets} to accelerate machine learning research on dietary sensing.

This paper presents several contributions as follows:

\begin{enumerate}
    \item Introduction of two novel food image datasets, namely NutritionVerse-Synth (NV-Synth) and NutritionVerse-Real (NV-Real), enriched with both diet information and segmentation masks.
    \item Evaluation of two approaches (indirect and direct prediction) for food estimation on the identical dataset, aiming to identify the most effective approach.
    \item Exploration of the benefits of incorporating depth information in food estimation tasks, accompanied by comprehensive experimental results.
    \item Valuable insights into the synergistic utilization of synthetic and real data to enhance the accuracy of diet estimation methods.
\end{enumerate}

\section{Related Work}
A number of prior works have explored computer vision techniques for food recognition and dietary intake estimation, though significant limitations persist in terms of scope, data, and methodology. Recently released quality food image datasets such as UECF Food 100~\cite{uecfood-100-dataset}, FoodX-251~\cite{foodx-251}, and Food2K~\cite{food2k-dataset} contain a significant number of food images with diverse food items. Unfortunately, the dietary information linked to these 2D images is not made available, posing a challenge in utilizing these datasets to estimate energy, macronutrient and micronutrient intake. In addition, existing datasets comprise of 2D images with fixed or randomly selected camera views that are discretely sampled~\cite{food-recog-promise, foodx-251, uecfood-100-dataset, food2k-dataset, chinesefoodnet, food-101}. These set views introduce bias in terms of how individuals take images with their camera which would affect the training and accuracy of the model. Recipe-related datasets, like Recipe1M~\cite{marin2019learning,salvador2017learning}, are extensively utilized in food recognition and recipe generation studies. However, these datasets lack crucial components such as food segmentation and ingredient labels, which make it very difficult for the task of estimating nutritional information. Chen and Ngo investigated a deep learning-based ingredient recognition system for cooking recipe retrieval and couples the problem of food categorization with ingredient recognition by simultaneously training the model with both tasks~\cite{10.1145/2964284.2964315}. Notably, their model does not examine the accuracy of food intake estimation and the images in their dataset only had an average of three recognizable ingredients, unrealistic of real-world scenarios~\cite{10.1145/2964284.2964315}. 

Bolaños and Radeva~\cite{7900117} proposed a method using the modified GoogLeNet architecture to simultaneously recognize and localize foods in images but did not estimate food volume or dietary information. DepthCalorieCam~\cite{10.1145/3347448.3357172} utilized visual-inertial odometry on a smartphone to estimate food volume and derive caloric content by multiplying the calories density of the food's category with the estimated size of the food. However, their contribution was only demonstrated on three food types. Menu-Match~\cite{7045971} provides an automated computer vision system for food logging and tracking of calories, but focus only on the restaurant scenario and have only 646 images in their dataset.  Comparable studies~\cite{7410503, 6748066} focus on recognizing meal contents and estimating calories from individual meals. However, the methodologies in~\cite{7410503} are also primarily tailored to restaurant scenarios, and there is limited testing conducted in settings outside of restaurants. On the other hand, the dataset and methodologies in~\cite{6748066} are not publicly available and are limited to only 8 food categories. Furthermore, all these works~\cite{10.1145/3347448.3357172,7045971,7410503,6748066} are constrained to calories, and do not predict other dietary components.

Nutrition5k~\cite{thames2021nutrition5k} presents a promising development in image-based recognition systems. However, a major limitation of the dataset is that the models are trained on images captured from only four specific viewpoints~\cite{thames2021nutrition5k}. This narrow range of viewpoints does not accurately reflect the diverse angles from which individuals typically capture meal images, limiting the model's ability to generalize to various real-life scenarios. Liang and Li~\cite{2017arXiv170507632L} also present a promising computer vision-based food calories estimation dataset and method, but their dataset is limited to only calories and includes only 2978 images~\cite{2017arXiv170507632L}. Furthermore, they require that images are taken with a specific calibration reference to ensure accurate calories estimation, infeasible for real-world usage~\cite{2017arXiv170507632L}. Table~\ref{tab:research-overview} provides a general overview of existing dietary intake estimation datasets and methods. As seen, NV-Synth and NV-Real datasets are the only ones that are publicly available and have annotation data (e.g., segmentation masks), and dietary information.

\section{Data Collection}
\subsection{NutritionVerse-Synth (NV-Synth)}
\begin{figure}[t]
    \centering
    \subfloat[Angle 1]{\includegraphics[width=.48\linewidth]{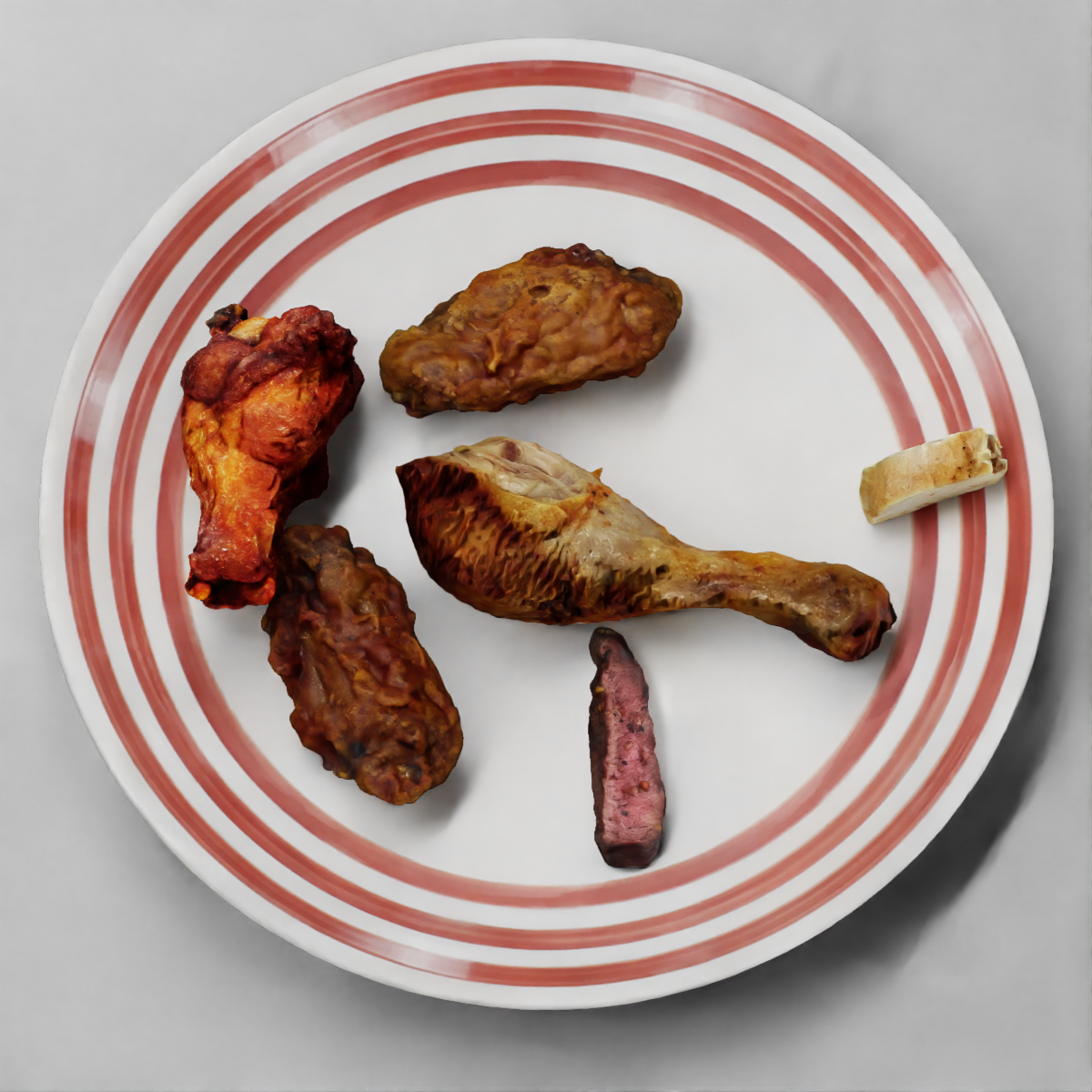}}
    \hfil
    \subfloat[Angle 2]{\includegraphics[width=.48\linewidth]{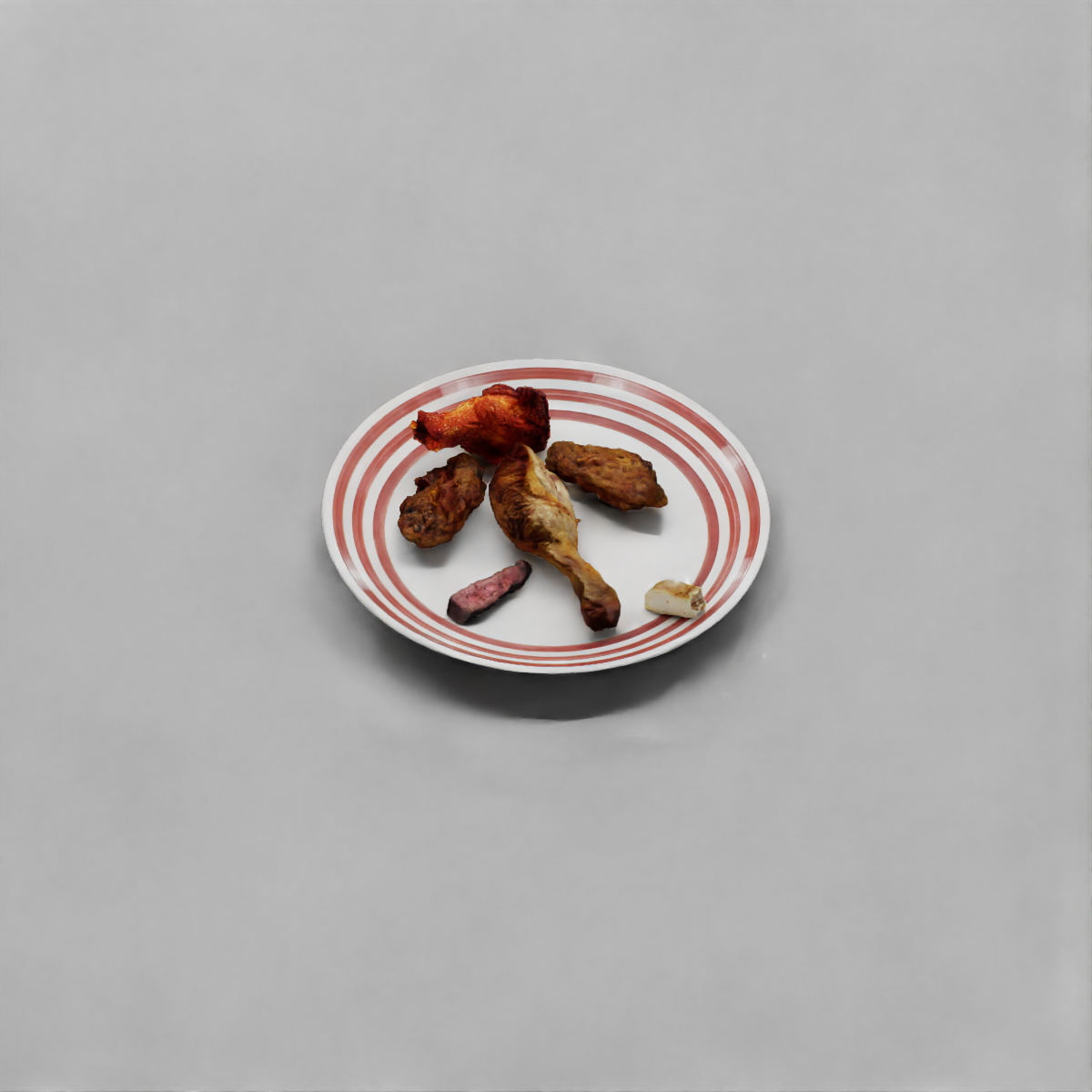}}
    \caption{An example food scene from NV-Synth with two different camera angles.}
    \label{fig:scene-2-camera-angles}
\end{figure}

Using the 3D meshes from the open access NutritionVerse-3D dataset~\cite{nutritionverse-3d}, Nvidia's Omniverse IsaacSim simulation framework~\cite{issac-sim} was used to generate synthetic scenes of meals. For each scene, up to 7 ingredients were sampled and then procedurally dropped onto a plate to simulate realistic food scenes. Using more than 7 often leads to items falling off the plate due to simulation physics. To maximize the realism and capture diverse plating conditions (including scenarios where the ingredients are highly disordered), the internal physics engine was leveraged to simulate physics-based interactions between ingredients of different masses and densities. Furthermore, realistic images were captured (e.g., some parts of dish are out of focus or occluded by other items) by using a variety of diverse and realistic camera perspectives and lighting conditions. The RGB image, corresponding depth image, associated object detection bounding boxes and segmentation masks were then generated using Omniverse for each scene for 12 random camera angles. An example of two random camera angles for a food scene is shown in Figure~\ref{fig:scene-2-camera-angles}. The nutritional metadata for the synthetic scenes was then calculated based on the metadata available in the NutritionVerse-3D dataset~\cite{nutritionverse-3d} and the outputted annotation metadata from Omniverse.

NV-Synth is a collection of 84,984 2D images of 7,082 distinct dishes with associated dietary metadata including mass, calories, carbohydrates, fats, and protein contents and ingredient labels for which food items are in each dish. 105 individual food items are represented in the dataset (with 45 unique food types), and the mean number of times each food item appeared in a food scene is 369.59. An average of 5.62 food items are present in each dish, and the mean dietary content of each food scene is 602.1 kcal, 315.9 g, 55.1 g, 34.2 g, and 30.0 g for calories, mass, protein, carbohydrate, and fat content, respectively. A subset of this dataset (28,328) was used for model development and was created by randomly selecting 4 different viewpoints (from 12 different angles) for each food scene. We use a 60\%/20\%/20\% training/validation/testing split of the scenes for the experiments and ensured all images from the same scene are kept in the same split. 

\subsection{NutritionVerse-Real (NV-Real)}
\begin{figure}[t]
    \centering
    \subfloat[Angle 1]{\includegraphics[width=.48\linewidth]{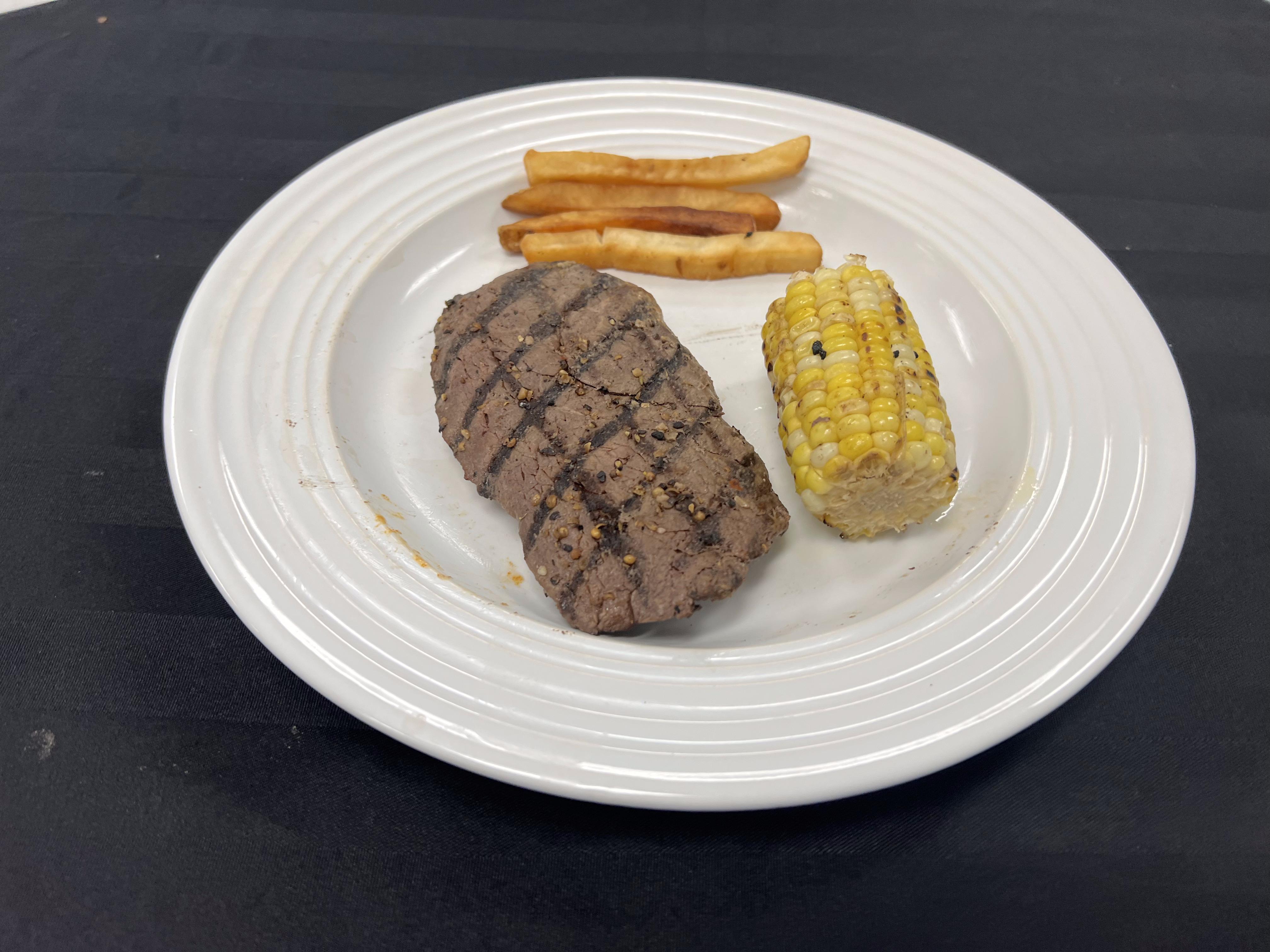}}
    \hfil
    \subfloat[Angle 2]{\includegraphics[width=.48\linewidth]{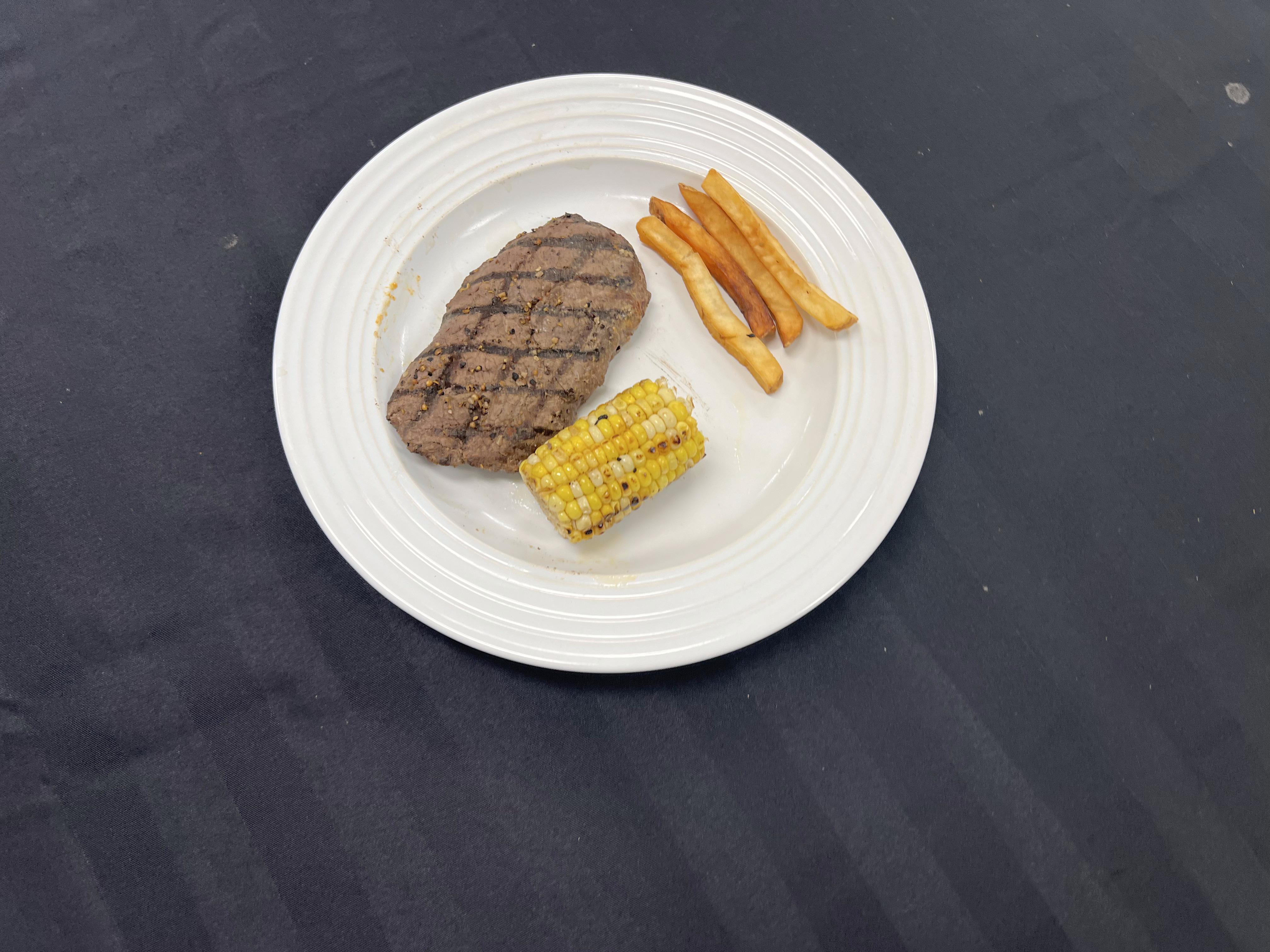}}
    \caption{An example food scene from NV-Real with two different camera angles.}
    \label{fig:real-scene-2-camera-angles}
\end{figure}

The NV-Real dataset was created by manually collecting images of food scenes in real life. The food items in the dishes was limited to those available in NutritionVerse-3D~\cite{nutritionverse-3d} to ensure appropriate verification of the approach. We used an iPhone 13 Pro Max~\cite{apple-iphone} to collect 10 images at random camera angles for each food dish. An example of two random camera angles for a food scene is shown in Figure~\ref{fig:real-scene-2-camera-angles}. To determine the dietary content of the dish, we measured the weight of every ingredient using a food scale. We then gathered the food composition information either from the packaging of the ingredients or from the Canada Nutrient File available on the Government of Canada website~\cite{canada-nutrient-file} in cases where packaging did not contain the dietary data. The segmentation masks was then obtained through human labelling of the images. For feasibility, four randomly selected images per dish were included in the annotation set to be labelled. Any images found with labelling inconsistencies were subsequently removed. We spent a total of 60 hours collecting images and 40 hours annotating the images.

NV-Real includes 889 2D images of 251 distinct dishes comprised of the real food items used to generate synthetic images. The metadata associated with the real-world dataset includes the type of food for each item on the plate with 45 unique food types present in the dataset. Each food item appears at least once in a dish an average of 18.29 times. The mean values represented in the scenes comprising the real-world dataset for calories, mass, protein, carbohydrate, and fat content are 830.0 kcal, 406.3 g, 59.9 g, 38.2 g, and 64.0 g, respectively. We use a 70\%/30\% training/testing split for the experiments and ensured all images from the same scene are kept in the same split. No validation data was required for the experiments as we used the same model hyperparameters from the synthetic experimental model runs for comparison parity between the synthetic and real training results. 

\section{Examined Approaches}
As seen in Table~\ref{tab:approach-overview}, there are two main approaches for dietary assessment: indirect and direct prediction. Unlike direct prediction, indirect prediction correlates dietary intake with the pixel counts of food items in an image. To determine the pixel count, segmentation models are employed to identify the image pixels corresponding to food items or classes. The obtained pixel count is then used to establish the association with the dietary intake.

\begin{table}
     \centering
    \begin{tabular}{ll}
    \hline
    \textbf{Approach} & \textbf{Work} \\ 
    \hline
    Direct &~\cite{10.1145/3347448.3357172,7045971,7410503,2017arXiv170507632L,thames2021nutrition5k} \\
    Indirect &~\cite{6748066}
    \end{tabular}
    \caption{Overview of approaches studied in literature.}
    \label{tab:approach-overview}
\end{table}

\begin{figure}[t]
    \centering
    \subfloat[Instance segmentation]{\includegraphics[height=.28\linewidth]{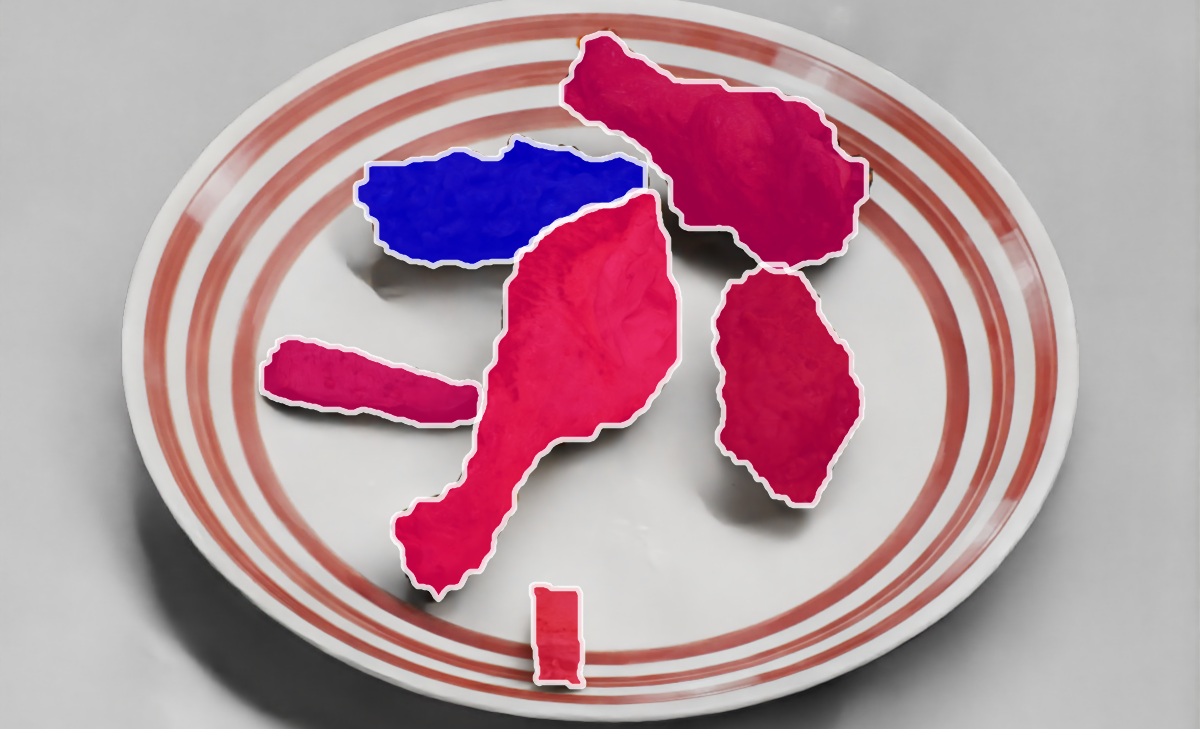}}
    \hfil
    \subfloat[Amodal instance segmentation]{\includegraphics[height=.28\linewidth]{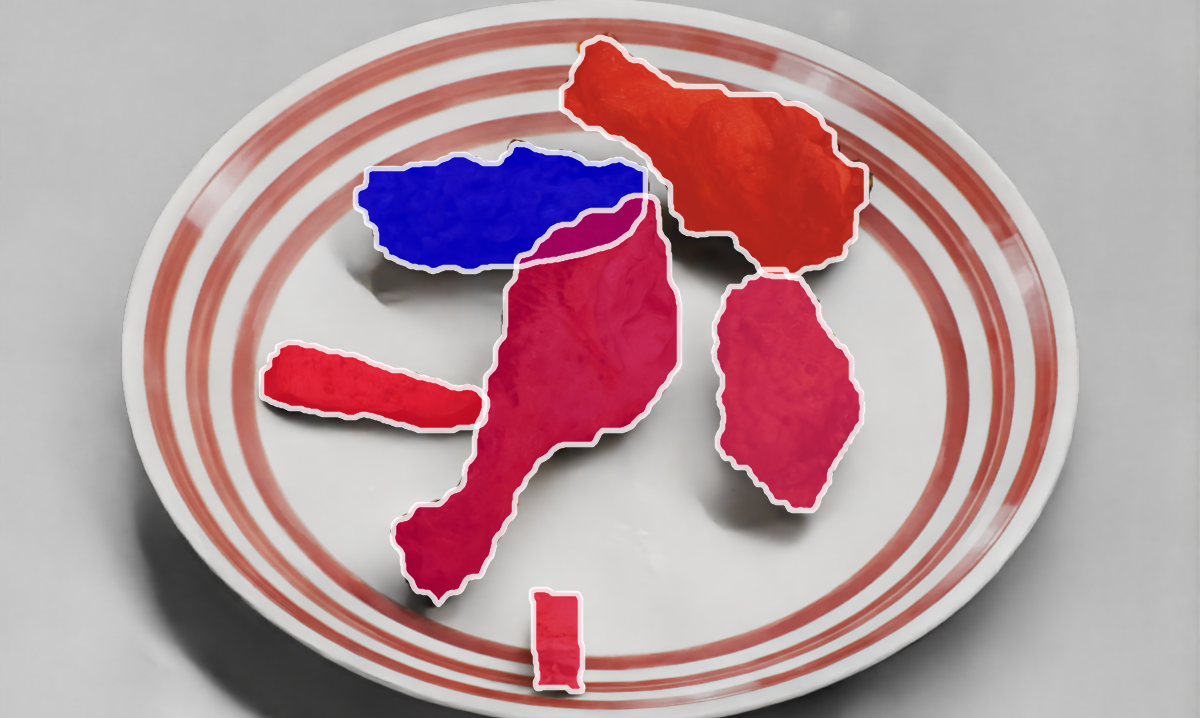}}
    \caption{The blue segmentation annotates a chicken wing that is partially occluded by a chicken leg in amodal instance compared to instance segmentation.}
    \label{fig:amodal_segmentation}
\end{figure}

There are three prominent types of segmentation models in literature: semantic, instance, and amodal instance segmentations. Semantic segmentation aims to classify each pixel in the image into predefined categories~\cite{cheng2022masked}. For dietary intake prediction, portion size is reflected in the number of pixels for each category. Instance segmentation extends beyond semantic segmentation and tries to also distinguish individual instances of objects, assigning unique labels to each pixel corresponding to a different object of the same category~\cite{he2017mask}. This is particularly useful when one of the instances is occluding the other, e.g., if there are two apples where one apple is partially occluded by the other but the model can identify that two apples exist in the dish. Amodal instance segmentation further builds on instance segmentation by accounting for occluded or partially obscured objects~\cite{back2022unseen} as seen in Figure~\ref{fig:amodal_segmentation}. By predicting a complete object mask, amodal helps under conditions where the object is heavily occluded, such as a burger buried in fries.

For direct prediction, various model architectures have been extensively studied in literature~\cite{10.1145/3347448.3357172,7045971,7410503,2017arXiv170507632L,thames2021nutrition5k} with the latest state-of-the-art model architecture being the Nutrition5k model architecture~\cite{thames2021nutrition5k} that estimates all five dietary intake tasks. 

\subsection{Model Hyperparameters}
\subsubsection{Direct Prediction}
Motivated by Nutrition5k~\cite{thames2021nutrition5k} which comprises an Inception-ResNet backbone encoder~\cite{DBLP:journals/corr/SzegedyVISW15} and a head module with four fully connected layers, we examine two deep learning architecture weight initializations to estimate the dietary information directly from the raw RGB image. For preprocessing, the RGB channels for the images were normalized based on their mean and standard deviation. We implemented the model architecture and hyperparameters used in the experimental setup for Nutrition5k~\cite{thames2021nutrition5k} and fine-tuned this architecture using two sets of pre-trained weights for the Inception-ResNet backbone encoder: (1) weights trained on the ImageNet dataset~\cite{DBLP:journals/corr/SzegedyVISW15} and (2) weights trained on the Nutrition5k dataset. These models were trained with 50 epochs and no early stopping criteria.

\subsubsection{Indirect Prediction}
Mask2Former~\cite{cheng2022masked}, Mask R-CNN~\cite{he2017mask}, and UOAIS-Net~\cite{back2022unseen} were used for prediction of semantic segmentation, instance segmentation, and amodal instance segmentation respectively. The original UOAIS-Net targeted at category-agnostic prediction of objects. Nevertheless, we found it to be effective in multi-category prediction when trained with multi-category ground truth labels.

For comparison parity, the same model hyperparameters were used for all experiments with the exception of base learning rate. A base learning rate of 0.02 was used for Mask R-CNN and UOAIS-Net which has similar architecture designed for instance segmentation. However, Mask2Former requires a lower base learning rate of 0.0001 to be stable because of its different architecture designed for semantic segmentation. We used SGD optimizer with momentum of 0.9, and weight decay of 0.0001 for training. The ResNet-50~\cite{he2016deep} backbone initialized with weights pretrained on ImageNet~\cite{russakovsky2015imagenet} was used for the indirect approach in all three segmentation methods. Those models were trained for 12 epochs using input size of 512x512 pixels and batch size of 16. 

\subsubsection{Depth Input}
Two model variations of each method were trained using 3-channel RGB input and 4-channel RGB-depth input respectively. The RGB channels were normalized based on their mean and standard deviation, and the depth channel was min-max normalized.

\subsection{Implementation Details}
\subsubsection{Direct Prediction}
Two weight initialization were considered for the backbone in the Nutrition5k direct prediction model architecture: weights from training on ImageNet~\cite{DBLP:journals/corr/SzegedyVISW15} and weights from training on the Nutrition5k dataset~\cite{thames2021nutrition5k} with normalized labels. The ImageNet weights were selected due to their widespread usage, while the Nutrition5k weights were used as Nutrition5k is state-of-the-art in food intake estimation. We report the performance for these two weight approaches as Direct Prediction (ImageNet) and Direct Prediction (Nutrition5k).

\subsubsection{Indirect Prediction}
Indirect prediction relies on assuming a linear relationship between the pixel count and the nutritional content specific to each food type. To establish this relationship, we leverage the data collected in the training set. For each nutrient, we estimate the average nutrient amount per pixel for each food type from the training set, using the ground truth data.

To obtain the pixel count, we follow a two-step process. First, we employ segmentation models to effectively segment the intake scene image, generating a segmentation mask for each food item. Second, we use these masks to count the number of pixels associated with each item.

By multiplying the pixel count with the average nutrient amount per pixel, we can effectively determine the dietary information for each nutrient and for each individual item in the scene. The comprehensive dietary intake information can then be derived by summing up all the nutrient values across all items within the scene. 

\begin{figure}[t]
    \centering
    \includegraphics[width=\linewidth]{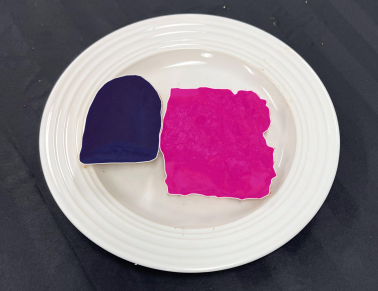}
    \caption{Example segmentation mask for a food dish with a half bread loaf (left) and lasagna (right) for nutrition calculation demonstration. The half bread loaf has a mask with 273,529 pixels, and the lasagna has a mask with 512,985 pixels.}
    \label{fig:example-mask-calculation}
\end{figure}

For example, Figure ~\ref{fig:example-mask-calculation} displays the example segmentation mask, depicting 273,529 pixels of the half bread loaf (left) and 512,985 pixels of lasagna (right). Given that the average calories per pixel for the half bread loaf is 9.08e-4 and the average calories for the lasagna is 6.36e-4, the total calories would equal:
\begin{center}
  273,529 * 9.08e-4 + 512,985 * 6.36e-4 = 574.3.  
\end{center}

Notably, the prediction results of semantic segmentation and instance segmentation are in different formats and requires different processing when calculating the pixel area. The semantic segmentation prediction result of each image is a mask where each pixel is assigned a label. The pixel area of each food ingredient can be counted without any preprocessing on the result. On the other hand, the instance segmentation prediction result of each image is a set of binary masks that each has an assigned label and a confidence score between 0 to 1 which represents the likelihood of correct prediction. Therefore, a threshold value needs to be chosen to filter out the predictions with low confidence scores. A parameter sweep for the threshold value in the range of 0 to 1 is conducted by applying the threshold filtering on all prediction results of the validation set and comparing the mean absolute error (MAE) for the five diet components. The threshold value that achieves the lowest MAE is chosen to be used on the processing of prediction results of the test set. Hence, it is possible that a different threshold value is chosen for the instance and amodal instance model using this method.

\section{Experiments}
The comprehensive datasets NV-Synth and NV-Real enable us to conduct novel experiments that are helpful in dietary assessment. Specifically, given the perfect labels in NV-Synth, we can evaluate different vision-based dietary assessment approaches to determine the most effective approach. We can also examine the merit of using depth information in dietary assessment. Depth directly relates to object volume and portion size which was shown to previously improve model performance~\cite{6748066,pfisterer2022automated}. Hence, we can compare the performance of models trained with and without depth information. Finally, being the pioneer in providing paired datasets comprising synthetic and real images, we can investigate the growing concern regarding the potential impact of synthetic data utilization on model performance in real-world scenarios. Notably, we can assess the synergistic utilization of synthetic and real data through three scenarios: (A) models trained solely on synthetic data, (B) models trained on synthetic data and fine-tuned on real data, and (C) models trained exclusively on real data, with the evaluation conducted on the on NV-Real test set. 

Notably, these three core questions are studied: 
\begin{enumerate}
    \item What is the best approach for dietary assessment? 
    \item Does depth information improve model performance?
    \item What is the impact of using synthetic data?
\end{enumerate}

\subsection{What is the best approach for dietary assessment?}
\begin{figure}[t]
    \centering
    \includegraphics[width=\linewidth]{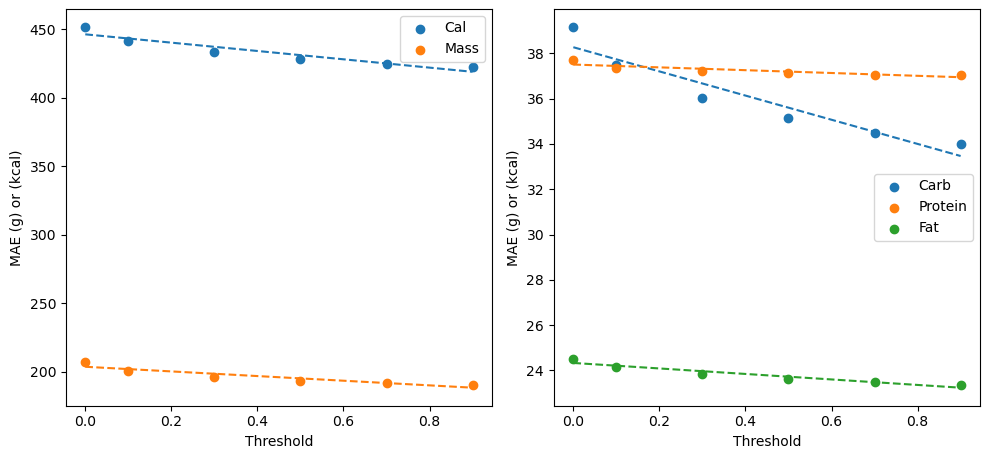}
    \caption{Validation MAE performance for the instance model for various confidence score thresholds.}
    \label{fig:thres-instance-mae}
\end{figure}

\begin{figure}[t]
    \centering
    \includegraphics[width=\linewidth]{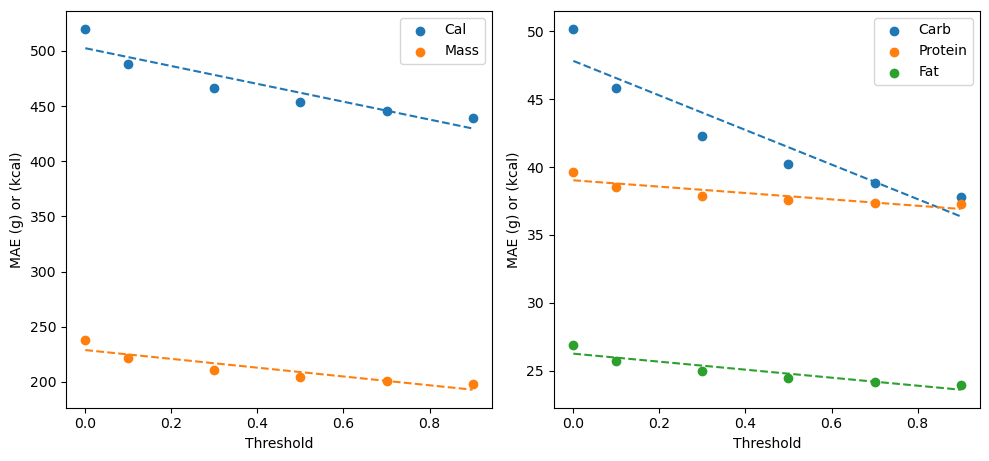}
    \caption{Validation MAE performance for the amodal instance model for various confidence score thresholds.}
    \label{fig:thres-amodal-mae}
\end{figure}

To answer this question, we compare the performance of the models trained using RGB images on the NV-Synth test set. As previously mentioned, for the indirect approach using the instance and amodal instance model, thresholding using the validation set was conducted. As seen in Figure~\ref{fig:thres-instance-mae} and Figure~\ref{fig:thres-amodal-mae}, the best threshold for both the instance and amodal instance model is 0.9 as it resulted in the best MAE values for the five diet components on the validation set. 

\begin{figure*}[t]
    \centering
    \subfloat[RGB input (Ground Truth) \\ CL: 1609, M: 684, P: 65, F: 69, CB: 183]{\includegraphics[width=.24\linewidth]{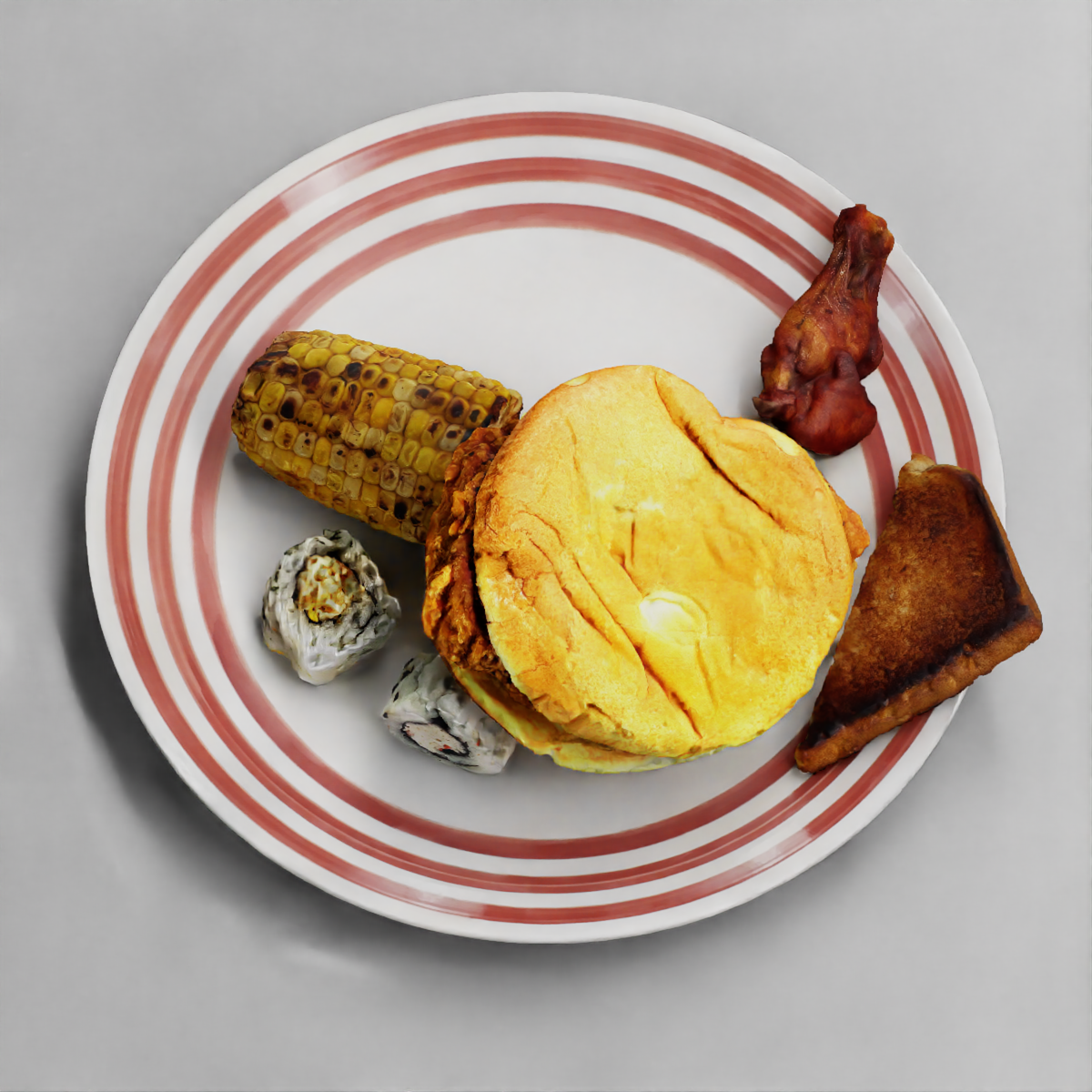}}
    \hfil
    \subfloat[Semantic segmentation \\ CL: 371, M: 156, P: 21, F: 18, CB: 32]{\includegraphics[width=.24\linewidth]{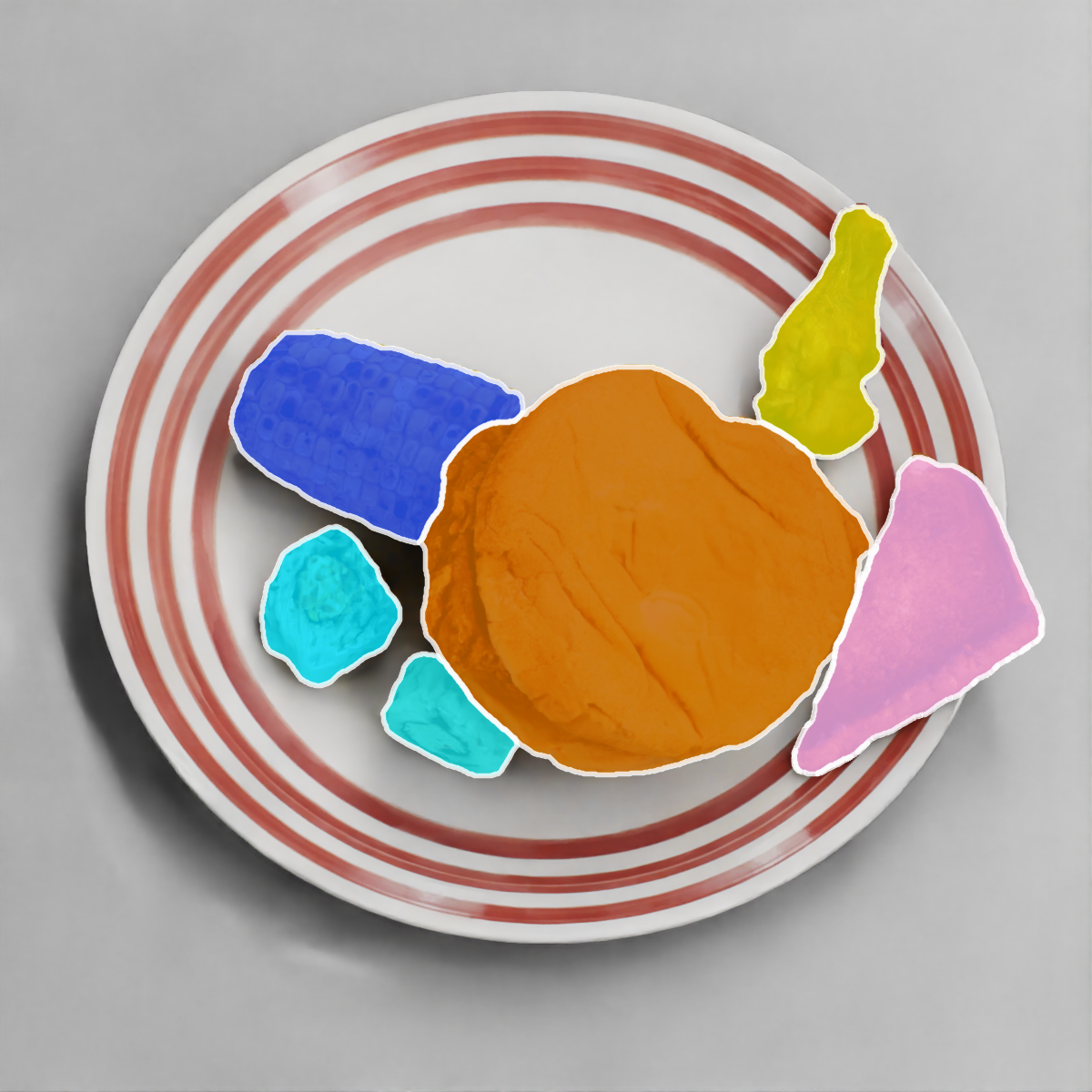}}
    \hfil
    \subfloat[Instance segmentation \\ CL: 706, M: 268, P: 39, F: 33, CB: 65]{\includegraphics[width=.24\linewidth]{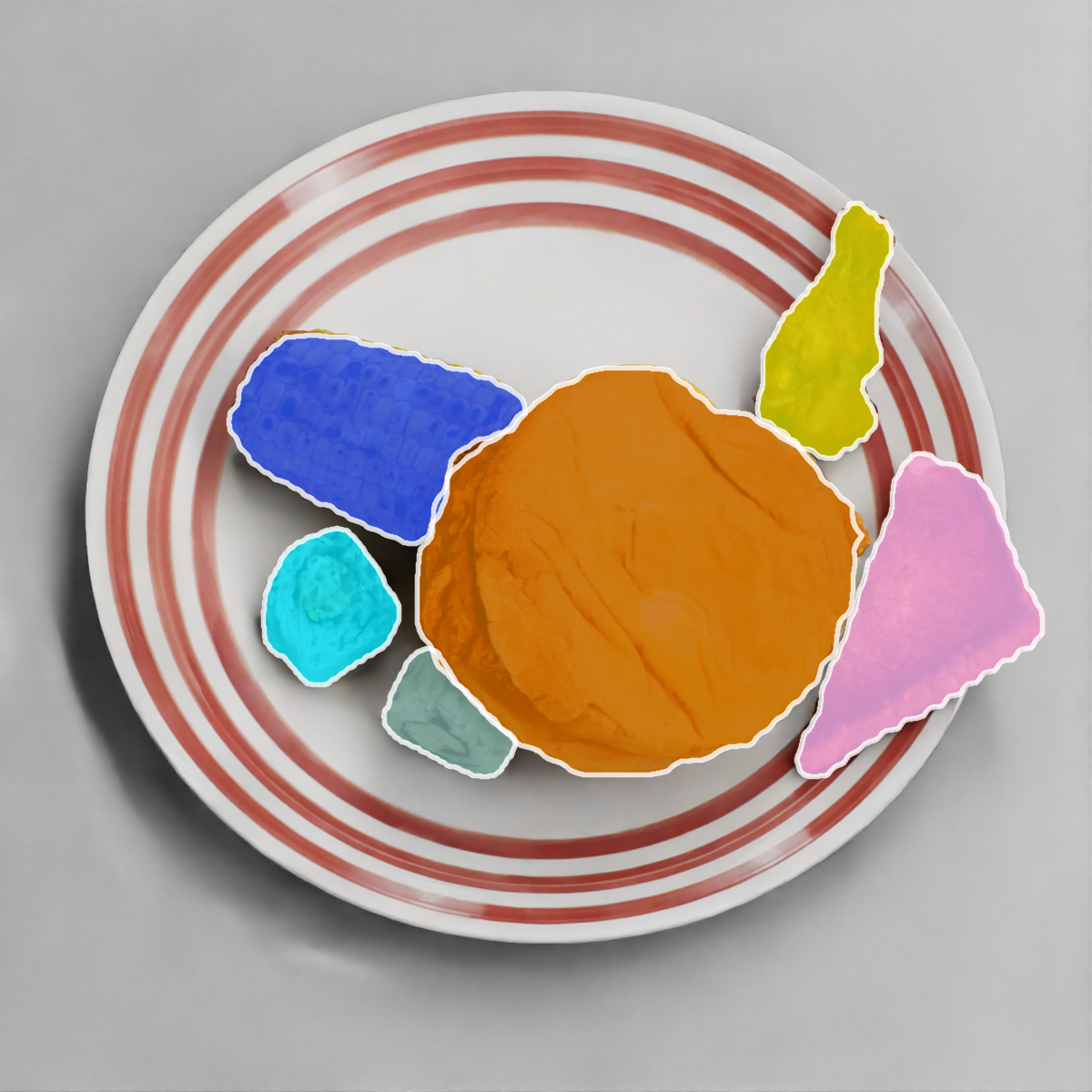}}
    \hfil
    \subfloat[Amodal instance segmentation \\ CL: 898, M: 337, P: 45, F: 40, CB: 90]{\includegraphics[width=.24\linewidth]{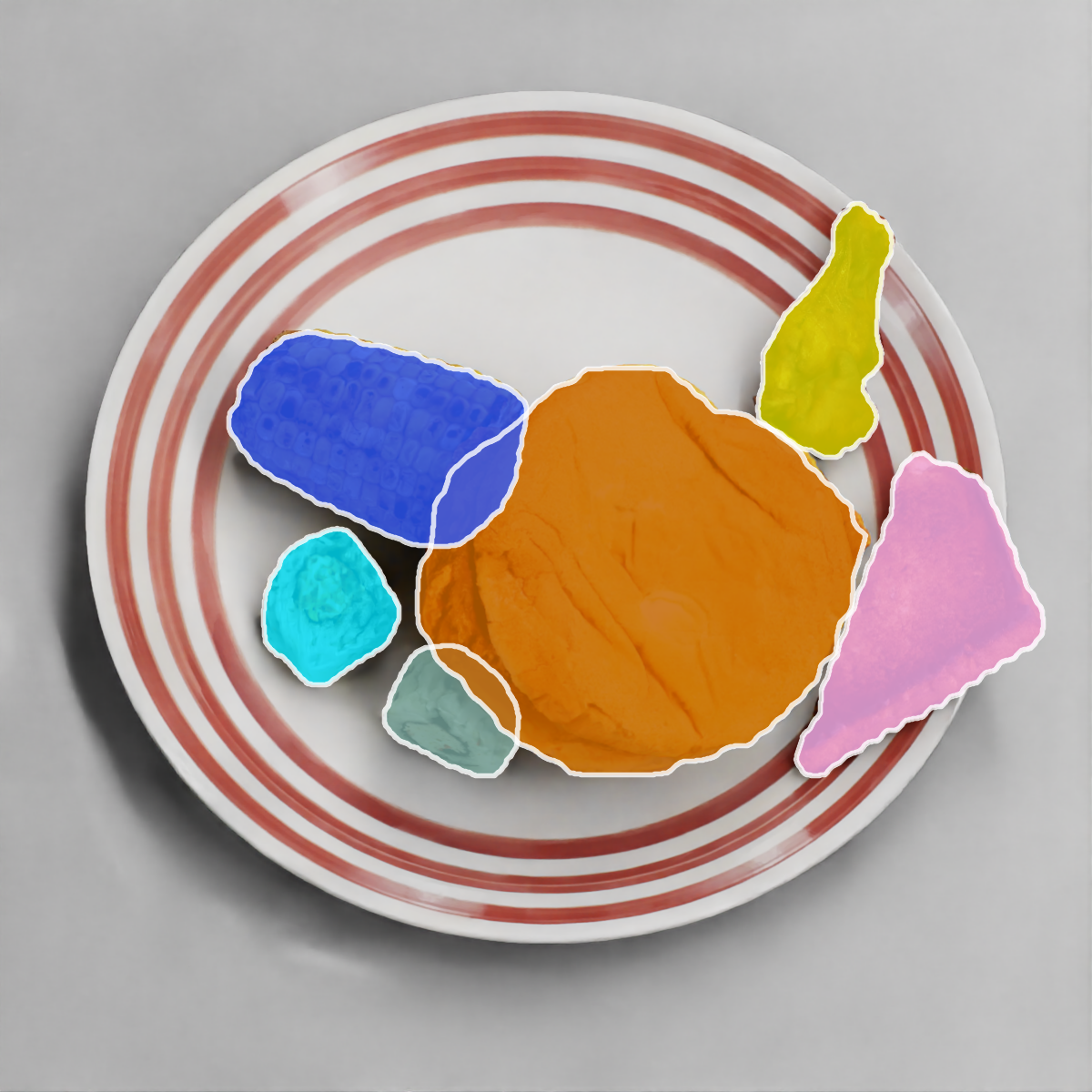}}
    \caption{Segmentation and prediction results of models trained with RGB input where CL refers to calories, M to mass, P to protein, F to fat, and CB to carbohydrate.}
    \label{fig:prediction_results}
\end{figure*}

\begin{table*}[!ht]
     \centering
    \begin{tabular}{llrrrrr}
    \hline
    \textbf{Model (RGB)} & \textbf{Eval Dataset} & \multicolumn{1}{l}{\textbf{Calories MAE}} & \multicolumn{1}{l}{\textbf{Mass MAE}} & \multicolumn{1}{l}{\textbf{Protein MAE}} & \multicolumn{1}{l}{\textbf{Fat MAE}} & \multicolumn{1}{l}{\textbf{Carb MAE}} \\
    \hline
    Semantic & NV-Synth & 418.1 & 185.4 & 39.0 & 23.5 & 32.3 \\
    Instance & NV-Synth & 430.9 & 191.4 & 39.3 & 24.1 & 34.4 \\
    Amodal Instance & NV-Synth & 451.3 & 202.8 & 39.6 & 24.8 & 38.5 \\
    \hline
    Direct Prediction (ImageNet) & NV-Synth & 229.2 & 102.6 & 56.0 & 12.0 & \textbf{19.4*} \\
    Direct Prediction (Nutrition5k) & NV-Synth & \textbf{128.7*} & \textbf{77.2*} & \textbf{18.5*} & \textbf{9.1*} & 21.5
    \end{tabular}
    \caption{Evaluation of model architectures using NV-Synth (RGB images) with the lowest MAE value for each column bolded with an * next to it.}
    \label{tab:synthetic-pred-results}
\end{table*}

Table~\ref{tab:synthetic-pred-results} shows the NV-Synth test set results for the model architectures trained on the NV-Synth train set with the lowest MAE for each nutrient bolded and indicated with an *. As seen in Table~\ref{tab:synthetic-pred-results}, semantic segmentation outperformed both instance and amodal instance methods for all dietary tasks with instance performing better than amodal instance. An example of the predicted segmentation masks and associated prediction results for the indirect approach is shown in Figure~\ref{fig:prediction_results}. Although the direct prediction models have the lowest MAE for at least one of the dietary components, Direct Prediction (Nutrition5k) performs the best holistically followed by Direct Prediction (ImageNet) as they generally have the lowest MAE across the five diet components. As such, the best approach for dietary assessment is using the direct prediction approach with initialization using Nutrition5k weights performing better than initialization using ImageNet weights.

\subsection{Does depth information improve model performance?}
\begin{table*}[!ht]
     \centering
    \begin{tabular}{llrrrrr}
    \hline
    \textbf{Model (RGBD)} & \textbf{Eval Dataset} & \multicolumn{1}{l}{\textbf{Calories MAE}} & \multicolumn{1}{l}{\textbf{Mass MAE}} & \multicolumn{1}{l}{\textbf{Protein MAE}} & \multicolumn{1}{l}{\textbf{Fat MAE}} & \multicolumn{1}{l}{\textbf{Carb MAE}} \\
    \hline
    Semantic & NV-Synth & 418.3 & 185.3 & 39.0 & 23.5 & 32.2 \\
    Instance & NV-Synth & 432.9 & 194.1 & 39.1 & 24.1 & 35.2 \\
    Amodal Instance & NV-Synth & 462.0 & 208.1 & 39.7 & 25.2 & 40.4 \\
    \hline
    Direct Prediction (ImageNet) & NV-Synth & 371.7 & 317.6 & 34.8 & \textbf{19.2*} & \textbf{25.2*} \\
    Direct Prediction (Nutrition5k) & NV-Synth & \textbf{202.0*} & \textbf{78.8*} & \textbf{23.5*} & 30.1 & 33.3
    \end{tabular}
    \caption{Investigation of depth information using NV-Synth (RGBD images) with the lowest MAE value for each column bolded with an * next to it.}
    \label{tab:synthetic-rgbd-pred-results}
\end{table*}

To answer this question, we compare the performance of the models with and without depth information using the NV-Synth test set. Table~\ref{tab:synthetic-rgbd-pred-results} shows the NV-Synth test set results for the model architectures trained on the RGBD images in the NV-Synth train set with the lowest MAE for each nutrient bolded and indicated with an *. As seen in Table~\ref{tab:synthetic-pred-results} and Table~\ref{tab:synthetic-rgbd-pred-results}, using depth for the direct prediction models leads to generally worse MAE values than using the pure RGB images, but using depth appears to improve the indirect approach with segmentation models. Hence, it appears that depth information does not improve model performance for direct prediction but may slightly help for the indirect approach. This finding is congruent with~\cite{thames2021nutrition5k} who observed a decline in their direct model performance when using depth images and~\cite{pfisterer2022automated} who observed an improvement with their indirect approach using segmentation models.

\subsection{What is the impact of using synthetic data?}
We investigate this question by comparing the performance on the NV-Real test set for three scenarios: (A) Using models trained only on NV-Synth, (B) Fine-tuning models trained on NV-Synth using NV-Real, and (C) Training models only on NV-Real. Notably, inference with the RGBD trained models and fine-tuning of the instance and amodal instance segmentation models is omitted due to the absence of depth and instance masks in the NV-Real dataset. 

\begin{table*}
     \centering
    \begin{tabular}{llrrrrr}
    \hline
    \textbf{Model (Scenario A)} & \textbf{Eval Dataset} & \multicolumn{1}{l}{\textbf{Calories MAE}} & \multicolumn{1}{l}{\textbf{Mass MAE}} & \multicolumn{1}{l}{\textbf{Protein MAE}} & \multicolumn{1}{l}{\textbf{Fat MAE}} & \multicolumn{1}{l}{\textbf{Carb MAE}} \\
    \hline
    Semantic & NV-Real & 40830.5 & 17342.0 & 2086.4 & 1630.4 & 4432.3 \\
    Instance & NV-Real & 50190.0 & 33774.6 & 2950.5 & 2009.5 & 5108.0 \\
    Amodal Instance & NV-Real & 72999.6 & 38379.2 & 4460.2 & 3225.3 & 6580.1 \\
    \hline
    Direct Prediction (ImageNet) & NV-Real & 530.6 & \textbf{182.9*} & 62.6 & 27.7 & \textbf{54.4*} \\
    Direct Prediction (Nutrition5k) & NV-Real & \textbf{525.9*} & 188.4 & \textbf{39.1*} & \textbf{27.4*} & 54.6
    \end{tabular}
    \caption{Scenario A: Models trained only on NV-Synth, with the lowest MAE value for each column bolded with an * next to it.}
    \label{tab:real-world-inference-results-synthetic-pretrained-only}
\end{table*}

\begin{table*}
     \centering
    \begin{tabular}{llrrrrr}
    \hline
    \textbf{Model (Scenario B)} & \textbf{Eval Dataset} & \textbf{Calories MAE} & \textbf{Mass MAE} & \textbf{Protein MAE} & \textbf{Fat MAE} & \textbf{Carb MAE} \\
    \hline
    Semantic & NV-Real & \textbf{445.1*} & 219.2 & 39.0 & \textbf{22.8*} & \textbf{41.8*} \\
    \hline
    Direct Prediction (ImageNet) & NV-Real & 475.7 & \textbf{149.1*} & 62.6 & 24.4 & 52.3 \\
    Direct Prediction (Nutrition5k) & NV-Real & 499.7 & 190.1 & \textbf{37.3*} & 25.4 & 53.9
    \end{tabular}
    \caption{Scenario B: Models trained on NV-Synth and fine-tuned on NV-Real, with the lowest MAE value for each column bolded with an * next to it.}
    \label{tab:real-world-inference-results-synthetic-finetuned}
\end{table*}

\begin{table*}
     \centering
    \begin{tabular}{llrrrrr}
    \hline
    \textbf{Model (Scenario C)} & \textbf{Eval Dataset} & \textbf{Calories MAE} & \textbf{Mass MAE} & \textbf{Protein MAE} & \textbf{Fat MAE} & \textbf{Carb MAE} \\
    \hline
    Semantic & NV-Real & 442.7 & 221.0 & \textbf{40.1*} & 23.0 & 42.4 \\
    \hline
    Direct Prediction (ImageNet) & NV-Real & \textbf{296.9*} & \textbf{115.9*} & 62.6 & \textbf{16.2*} & \textbf{29.8*} \\
    Direct Prediction (Nutrition5k) & NV-Real & 430.0 & 145.6 & 62.6 & 37.7 & 55.3
    \end{tabular}
    \caption{Scenario C: Models trained only on NV-Real, with the lowest MAE value for each column bolded with an * next to it.}
    \label{tab:real-world-inference-results-real-trained}
\end{table*}

\begin{table*}
     \centering
    \begin{tabular}{lllrrrrr}
    \hline
    \textbf{Model Description} & \textbf{Trained} & \textbf{Fine-Tuned} & \textbf{Calories MAE} & \textbf{Mass MAE} & \textbf{Protein MAE} & \textbf{Fat MAE} & \multicolumn{1}{l}{\textbf{Carb MAE}} \\
    \hline
    (A) Direct Prediction (Nutrition5k) & NV-Synth & N/A & 525.9 & 188.4 & 39.1 & 27.4 & 54.6 \\
    (B) Semantic & NV-Synth & NV-Real & 445.1 & 219.2 & \textbf{39.0*} & 22.8 & 41.8 \\
    (C) Direct Prediction (ImageNet) & NV-Real & N/A & \textbf{296.9*} & \textbf{115.9*} & 62.6 & \textbf{16.2*} & \textbf{29.8*}
    \end{tabular}
    \caption{Comparison of the best model from the three scenarios evaluated on the NV-Real dataset, with the lowest MAE value for each column bolded with an * next to it.}
    \label{tab:best-model-results}
\end{table*}

When looking at the model performance for models trained solely on the synthetic data (Scenario A), the direct prediction models have the lowest MAE for at least one of the dietary components and outperform the segmentation models as seen in Table~\ref{tab:real-world-inference-results-synthetic-pretrained-only}. Significantly higher MAE values were observed with the indirect approach employing segmentation models. This discrepancy can be attributed to the utilization of average pixel counts from the synthetic dataset, which do not align with the individual food items' average pixel counts in the real dataset. These variations stem from differences in camera setups during data collection and highlight an area of improvement for the synthetic dataset. The best model based on the lowest MAE across the five diet components was the Direct Prediction (Nutrition5k) model.

On the other hand, for the fine-tuned models (Scenario B), the Semantic model had generally better MAE performance than the other models except for mass and protein, where the Direct Prediction models achieved lower MAE values (Table~\ref{tab:real-world-inference-results-synthetic-finetuned}). Fine-tuning the model generally resulted in better (lower) MAE values for the Semantic and Direct Prediction models.

For models trained exclusively on real data, the results in Table~\ref{tab:real-world-inference-results-real-trained} shows that the Direct Prediction model (initialized with ImageNet weights)  trained on the NV-Real train set has the lowest MAE for most of the dietary components (except for protein) compared to the other models trained on the NV-Real train set. 

Through the comparisons, the best model (deemed by the lowest MAE for NV-Real test set) generally across the five diet components is the Direct Prediction (ImageNet) model trained on the NV-Real train set (as seen in Table~\ref{tab:best-model-results}). Notably, the semantic model trained on NV-Synth and fine-tuned on NV-Real achieves better performance for protein but the semantic model has higher MAE scores for the other four dietary components compared to the Direct Prediction (ImageNet) model trained on NV-Real.

\section{Conclusion}
In this paper, we investigate various intake estimation approaches and introduce two new food datasets with associated food composition information: NV-Synth (created using the open access NV-3D dataset) and NV-Real (manually collected). Unlike other datasets, NV-Synth contains a comprehensive set of labels that no other dataset has, including depth images, instance masks, and semantic masks. With these comprehensive labels, we compared various approaches side-by-side to determine the best approach for dietary estimation. We then attempted to verify our findings using the NV-Real dataset and found that the Direct Prediction (ImageNet)  model trained on the NV-Real dataset achieves the best performance. Interestingly, it was more advantageous to leverage the weights trained on the ImageNet dataset rather than the weights trained on the Nutrition5k dataset. Hence, our results indicate that it is still advantageous to train on the real images rather than leverage synthetic images for model training. Future work involves iterating on the synthetic dataset to more closely mirror images collected in real life through increasing the diversity of images and viewpoints per scene and applying these models on an external food dataset to validate their generalization to different situations. 

\section*{Acknowledgements}
This work was supported by the National Research Council Canada (NRC) through the Aging in Place (AiP) Challenge Program, project number AiP-006. The authors also thank the graduate student partner in the Kinesiology and Health Sciences department Meagan Jackson and undergraduate research assistants Tanisha Nigam, Komal Vachhani, and Cosmo Zhao.

\bibliographystyle{IEEEtran}
\bibliography{refs}

% Generated by IEEEtran.bst, version: 1.14 (2015/08/26)
\begin{thebibliography}{10}
\providecommand{\url}[1]{#1}
\csname url@samestyle\endcsname
\providecommand{\newblock}{\relax}
\providecommand{\bibinfo}[2]{#2}
\providecommand{\BIBentrySTDinterwordspacing}{\spaceskip=0pt\relax}
\providecommand{\BIBentryALTinterwordstretchfactor}{4}
\providecommand{\BIBentryALTinterwordspacing}{\spaceskip=\fontdimen2\font plus
\BIBentryALTinterwordstretchfactor\fontdimen3\font minus \fontdimen4\font\relax}
\providecommand{\BIBforeignlanguage}[2]{{%
\expandafter\ifx\csname l@#1\endcsname\relax
\typeout{** WARNING: IEEEtran.bst: No hyphenation pattern has been}%
\typeout{** loaded for the language `#1'. Using the pattern for}%
\typeout{** the default language instead.}%
\else
\language=\csname l@#1\endcsname
\fi
#2}}
\providecommand{\BIBdecl}{\relax}
\BIBdecl

\bibitem{nutritionverse-3d}
\BIBentryALTinterwordspacing
C.-e.~A. Tai, M.~Keller, M.~Kerrigan, Y.~Chen, S.~Nair, P.~Xi, and A.~Wong, ``Nutritionverse-3d: A 3d food model dataset for nutritional intake estimation,'' in \emph{Conference on Computer Vision and Pattern Recognition (CVPR)}, ser. Women in Computer Vision (WiCV).\hskip 1em plus 0.5em minus 0.4em\relax Vancouver: IEEE, 2023. [Online]. Available: \url{https://arxiv.org/abs/2304.05619}
\BIBentrySTDinterwordspacing

\bibitem{10.1145/3347448.3357172}
\BIBentryALTinterwordspacing
Y.~Ando, T.~Ege, J.~Cho, and K.~Yanai, ``Depthcaloriecam: A mobile application for volume-based foodcalorie estimation using depth cameras,'' in \emph{Proceedings of the 5th International Workshop on Multimedia Assisted Dietary Management}, ser. MADiMa '19.\hskip 1em plus 0.5em minus 0.4em\relax New York, NY, USA: Association for Computing Machinery, 2019, p. 76–81. [Online]. Available: \url{https://doi.org/10.1145/3347448.3357172}
\BIBentrySTDinterwordspacing

\bibitem{7045971}
O.~Beijbom, N.~Joshi, D.~Morris, S.~Saponas, and S.~Khullar, ``Menu-match: Restaurant-specific food logging from images,'' in \emph{2015 IEEE Winter Conference on Applications of Computer Vision}.\hskip 1em plus 0.5em minus 0.4em\relax Waikoloa: IEEE, 2015, pp. 844--851.

\bibitem{7410503}
A.~Myers, N.~Johnston, V.~Rathod, A.~Korattikara, A.~Gorban, N.~Silberman, S.~Guadarrama, G.~Papandreou, J.~Huang, and K.~Murphy, ``Im2calories: Towards an automated mobile vision food diary,'' in \emph{2015 IEEE International Conference on Computer Vision (ICCV)}.\hskip 1em plus 0.5em minus 0.4em\relax Santiago: IEEE, 2015, pp. 1233--1241.

\bibitem{2017arXiv170507632L}
Y.~Liang and J.~Li, ``Computer vision-based food calorie estimation: dataset, method, and experiment,'' \emph{arXiv e-prints}, vol. abs/1705.07632, p. arXiv:1705.07632, May 2017.

\bibitem{thames2021nutrition5k}
Q.~Thames, A.~Karpur, W.~Norris, F.~Xia, L.~Panait, T.~Weyand, and J.~Sim, ``Nutrition5k: Towards automatic nutritional understanding of generic food,'' in \emph{Proceedings of the IEEE/CVF Conference on Computer Vision and Pattern Recognition}.\hskip 1em plus 0.5em minus 0.4em\relax Nashville: IEEE, 2021, pp. 8903--8911.

\bibitem{6748066}
P.~Pouladzadeh, S.~Shirmohammadi, and R.~Al-Maghrabi, ``Measuring calorie and nutrition from food image,'' \emph{IEEE Transactions on Instrumentation and Measurement}, vol.~63, no.~8, pp. 1947--1956, 2014.

\bibitem{malnutrition-qol}
\BIBentryALTinterwordspacing
H.~H. Keller, T.~Østbye, and G.~Richard, ``Nutritional risk predicts quality of life in elderly community-living canadians,'' \emph{The Journals of Gerontology: Series A}, vol.~59, no.~1, p. M68–M74, 2004. [Online]. Available: \url{https://academic.oup.com/biomedgerontology/article/59/1/M68/533583}
\BIBentrySTDinterwordspacing

\bibitem{automated-dietary-recall}
\BIBentryALTinterwordspacing
A.~F. Subar, S.~I. Kirkpatrick, B.~Mittl, T.~P. Zimmerman, F.~E. Thompson, C.~Bingley, G.~Willis, N.~G. Islam, T.~Baranowski, S.~McNutt, and N.~Potischman, ``The automated self-administered 24-hour dietary recall (asa24): A resource for researchers, clinicians, and educators from the national cancer institute,'' \emph{Journal of the Academy of Nutrition and Dietetics}, vol. 112, no.~8, pp. 1134--1137, 2012. [Online]. Available: \url{https://www.sciencedirect.com/science/article/pii/S2212267212005898}
\BIBentrySTDinterwordspacing

\bibitem{kipnis2003structure}
V.~Kipnis, A.~F. Subar, D.~Midthune, L.~S. Freedman, R.~Ballard-Barbash, R.~P. Troiano, S.~Bingham, D.~A. Schoeller, A.~Schatzkin, and R.~J. Carroll, ``Structure of dietary measurement error: results of the open biomarker study,'' \emph{American journal of epidemiology}, vol. 158, no.~1, pp. 14--21, 2003.

\bibitem{freedman2014pooled}
L.~S. Freedman, J.~M. Commins, J.~E. Moler, L.~Arab, D.~J. Baer, V.~Kipnis, D.~Midthune, A.~J. Moshfegh, M.~L. Neuhouser, R.~L. Prentice \emph{et~al.}, ``Pooled results from 5 validation studies of dietary self-report instruments using recovery biomarkers for energy and protein intake,'' \emph{American journal of epidemiology}, vol. 180, no.~2, pp. 172--188, 2014.

\bibitem{freedman2015pooled}
L.~S. Freedman, J.~M. Commins, J.~E. Moler, W.~Willett, L.~F. Tinker, A.~F. Subar, D.~Spiegelman, D.~Rhodes, N.~Potischman, M.~L. Neuhouser \emph{et~al.}, ``Pooled results from 5 validation studies of dietary self-report instruments using recovery biomarkers for potassium and sodium intake,'' \emph{American journal of epidemiology}, vol. 181, no.~7, pp. 473--487, 2015.

\bibitem{mobile-phone-applications}
\BIBentryALTinterwordspacing
S.~P. Elbert, A.~Dijkstra, and A.~Oenema, ``A mobile phone app intervention targeting fruit and vegetable consumption: The efficacy of textual and auditory tailored health information tested in a randomized controlled trial,'' \emph{Journal of Medical Internet Research}, vol.~18, no.~6, p. e147, 2016. [Online]. Available: \url{https://www.jmir.org/2016/6/e147}
\BIBentrySTDinterwordspacing

\bibitem{snap-and-eat}
\BIBentryALTinterwordspacing
W.~Zhang, Q.~Yu, B.~Siddiquie, A.~Divakaran, and H.~Sawhney, ``“snap-n-eat”: Food recognition and nutrition estimation on a smartphone,'' \emph{Journal of Diabetes Science and Technology}, vol.~9, no.~3, pp. 525--533, 2015, pMID: 25901024. [Online]. Available: \url{https://doi.org/10.1177/1932296815582222}
\BIBentrySTDinterwordspacing

\bibitem{digital-photography}
\BIBentryALTinterwordspacing
D.~A. Williamson, H.~R. Allen, P.~D. Martin, A.~J. Alfonso, B.~Gerald, and A.~Hunt, ``Comparison of digital photography to weighed and visual estimation of portion sizes,'' \emph{Journal of the American Dietetic Association}, vol. 103, no.~9, p. 1139–1145, 2003. [Online]. Available: \url{https://www.sciencedirect.com/science/article/pii/S000282230300974X}
\BIBentrySTDinterwordspacing

\bibitem{personal-assistant}
\BIBentryALTinterwordspacing
A.~Rusu, M.~Randriambelonoro, C.~Perrin, C.~Valk, B.~Álvarez, and A.-K. Schwarze, ``Aspects influencing food intake and approaches towards personalising nutrition in the elderly,'' \emph{Journal of Population Ageing}, vol.~13, p. 239–256, 2020. [Online]. Available: \url{https://link.springer.com/article/10.1007/s12062-019-09259-1}
\BIBentrySTDinterwordspacing

\bibitem{food-recog-promise}
G.~Ciocca, P.~Napoletano, and R.~Schettini, ``Food recognition: A new dataset, experiments, and results,'' \emph{IEEE Journal of Biomedical and Health Informatics}, vol.~21, no.~3, pp. 588--598, 2017.

\bibitem{uecfood-100-dataset}
Y.~Matsuda, H.~Hoashi, and K.~Yanai, ``Recognition of multiple-food images by detecting candidate regions,'' in \emph{2012 IEEE International Conference on Multimedia and Expo}.\hskip 1em plus 0.5em minus 0.4em\relax Melbourne: IEEE, 2012, pp. 25--30.

\bibitem{foodx-251}
\BIBentryALTinterwordspacing
P.~Kaur, K.~Sikka, W.~Wang, S.~J. Belongie, and A.~Divakaran, ``Foodx-251: {A} dataset for fine-grained food classification,'' \emph{CoRR}, vol. abs/1907.06167, 2019. [Online]. Available: \url{http://arxiv.org/abs/1907.06167}
\BIBentrySTDinterwordspacing

\bibitem{food2k-dataset}
\BIBentryALTinterwordspacing
W.~Min, Z.~Wang, Y.~Liu, M.~Luo, L.~Kang, X.~Wei, X.~Wei, and S.~Jiang, ``Large scale visual food recognition,'' \emph{CoRR}, vol. abs/2103.16107, 2021. [Online]. Available: \url{https://arxiv.org/abs/2103.16107}
\BIBentrySTDinterwordspacing

\bibitem{chinesefoodnet}
\BIBentryALTinterwordspacing
X.~Chen, H.~Zhou, and L.~Diao, ``Chinesefoodnet: {A} large-scale image dataset for chinese food recognition,'' \emph{CoRR}, vol. abs/1705.02743, 2017. [Online]. Available: \url{http://arxiv.org/abs/1705.02743}
\BIBentrySTDinterwordspacing

\bibitem{food-101}
L.~Bossard, M.~Guillaumin, and L.~Van~Gool, ``Food-101 -- mining discriminative components with random forests,'' in \emph{European Conference on Computer Vision}.\hskip 1em plus 0.5em minus 0.4em\relax Zurich: Springer Science and Business Media, 2014, p. 446–461.

\bibitem{marin2019learning}
J.~Marin, A.~Biswas, F.~Ofli, N.~Hynes, A.~Salvador, Y.~Aytar, I.~Weber, and A.~Torralba, ``Recipe1m+: A dataset for learning cross-modal embeddings for cooking recipes and food images,'' \emph{IEEE Transactions on Pattern Analysis and Machine Intelligence}, vol.~43, no.~1, p. 187–203, 2019.

\bibitem{salvador2017learning}
A.~Salvador, N.~Hynes, Y.~Aytar, J.~Marin, F.~Ofli, I.~Weber, and A.~Torralba, ``Learning cross-modal embeddings for cooking recipes and food images,'' in \emph{2017 IEEE Conference on Computer Vision and Pattern Recognition (CVPR)}.\hskip 1em plus 0.5em minus 0.4em\relax Honolulu: IEEE, 2017, pp. 3068--3076.

\bibitem{10.1145/2964284.2964315}
\BIBentryALTinterwordspacing
J.~Chen and C.-w. Ngo, ``Deep-based ingredient recognition for cooking recipe retrieval,'' in \emph{Proceedings of the 24th ACM International Conference on Multimedia}, ser. MM '16.\hskip 1em plus 0.5em minus 0.4em\relax New York, NY, USA: Association for Computing Machinery, 2016, p. 32–41. [Online]. Available: \url{https://doi.org/10.1145/2964284.2964315}
\BIBentrySTDinterwordspacing

\bibitem{7900117}
M.~Bolaños and P.~Radeva, ``Simultaneous food localization and recognition,'' in \emph{2016 23rd International Conference on Pattern Recognition (ICPR)}.\hskip 1em plus 0.5em minus 0.4em\relax Cancun: IEEE, 2016, pp. 3140--3145.

\bibitem{issac-sim}
\BIBentryALTinterwordspacing
NVIDIA. (2023) Nvidia isaac sim. NVIDIA. [Online]. Available: \url{https://developer.nvidia.com/isaac-sim}
\BIBentrySTDinterwordspacing

\bibitem{apple-iphone}
\BIBentryALTinterwordspacing
Apple. (2022) iphone. Apple. [Online]. Available: \url{https://www.apple.com/ca/iphone/}
\BIBentrySTDinterwordspacing

\bibitem{canada-nutrient-file}
\BIBentryALTinterwordspacing
G.~of~Canada, ``Canadian nutrient file (cnf) - search by food,'' 2022. [Online]. Available: \url{https://food-nutrition.canada.ca/cnf-fce/}
\BIBentrySTDinterwordspacing

\bibitem{cheng2022masked}
B.~Cheng, I.~Misra, A.~G. Schwing, A.~Kirillov, and R.~Girdhar, ``Masked-attention mask transformer for universal image segmentation,'' in \emph{Proceedings of the IEEE/CVF conference on computer vision and pattern recognition}.\hskip 1em plus 0.5em minus 0.4em\relax New Orleans: IEEE, 2022, pp. 1290--1299.

\bibitem{he2017mask}
K.~He, G.~Gkioxari, P.~Doll{\'a}r, and R.~Girshick, ``Mask r-cnn,'' in \emph{Proceedings of the IEEE international conference on computer vision}.\hskip 1em plus 0.5em minus 0.4em\relax Venice: IEEE, 2017, pp. 2961--2969.

\bibitem{back2022unseen}
S.~Back, J.~Lee, T.~Kim, S.~Noh, R.~Kang, S.~Bak, and K.~Lee, ``Unseen object amodal instance segmentation via hierarchical occlusion modeling,'' in \emph{2022 International Conference on Robotics and Automation (ICRA)}, IEEE.\hskip 1em plus 0.5em minus 0.4em\relax Philadelphia: IEEE, 2022, pp. 5085--5092.

\bibitem{DBLP:journals/corr/SzegedyVISW15}
C.~Szegedy, S.~Ioffe, V.~Vanhoucke, and A.~Alemi, ``Inception-v4, inception-resnet and the impact of residual connections on learning,'' 2016.

\bibitem{he2016deep}
K.~He, X.~Zhang, S.~Ren, and J.~Sun, ``Deep residual learning for image recognition,'' in \emph{2016 IEEE Conference on Computer Vision and Pattern Recognition (CVPR)}.\hskip 1em plus 0.5em minus 0.4em\relax Las Vegas: IEEE, 2016, pp. 770--778.

\bibitem{russakovsky2015imagenet}
O.~Russakovsky, J.~Deng, H.~Su, J.~Krause, S.~Satheesh, S.~Ma, Z.~Huang, A.~Karpathy, A.~Khosla, M.~Bernstein, A.~C. Berg, and L.~Fei-Fei, ``Imagenet large scale visual recognition challenge,'' \emph{Int. J. Comput. Vis.}, vol. 115, no.~3, pp. 211--252, Dec. 2015.

\bibitem{pfisterer2022automated}
K.~J. Pfisterer, R.~Amelard, A.~G. Chung, B.~Syrnyk, A.~MacLean, H.~H. Keller, and A.~Wong, ``Automated food intake tracking requires depth-refined semantic segmentation to rectify visual-volume discordance in long-term care homes,'' \emph{Scientific reports}, vol.~12, no.~1, p.~83, 2022.

\end{thebibliography}

\end{document}